\documentclass[10pt,journal,compsoc]{IEEEtran}
\usepackage{varwidth}
\usepackage[usenames,dvipsnames,svgnames,table]{xcolor}

\makeatletter
\def\mathcolor#1#{\@mathcolor{#1}}
\def\@mathcolor#1#2#3{%
  \protect\leavevmode
  \begingroup
    \color#1{#2}#3%
  \endgroup
}
\makeatother  

\usepackage{hyperref}
\hypersetup{
    colorlinks=true,
    linkcolor=blue,
    filecolor=magenta,      
    urlcolor=blue,
}

\usepackage{enumitem,amssymb}

\usepackage{hyperref}
\usepackage{booktabs}
\usepackage{subfig}

\usepackage{graphicx}
\usepackage{amsmath}
\usepackage{algorithm}
\usepackage{algorithmic}

\ifCLASSOPTIONcompsoc
  \usepackage[nocompress]{cite}
\else
  \usepackage{cite}
\fi
\ifCLASSINFOpdf
\else
\fi
\begin{document}
%
\title{CIZSL++: Creativity Inspired  Generative Zero-Shot Learning}
%
%
%
%

\author{Mohamed Elhoseiny, Kai Yi, 
        Mohamed Elfeki\\
\thanks{Manuscript received date TBD.}}  

%
%

\markboth{IEEE TRANSACTIONS ON PATTERN ANALYSIS AND MACHINE INTELLIGENCE}%
{Shell \MakeLowercase{\textit{et al.}}: Bare Demo of IEEEtran.cls for Computer Society Journals}
%



\IEEEtitleabstractindextext{%
\begin{abstract}
Zero-shot learning (ZSL) aims at understanding unseen categories with no training examples from class-level descriptions. To improve the discriminative power of ZSL, we model the visual learning process of unseen categories with inspiration from the psychology of human creativity for producing novel art. First, we propose CIZSL-v1 as a creativity inspired model for generative ZSL. We relate ZSL to human creativity by observing that ZSL is about recognizing the unseen, and creativity is about creating a likable unseen. We introduce a learning signal inspired by creativity literature that explores the unseen space with hallucinated class-descriptions and encourages careful deviation of their visual feature generations from seen classes while allowing knowledge transfer from seen to unseen classes. Second,  CIZSL-v2 is proposed as an improved version of CIZSL-v1 for generative zero-shot learning. CIZSL-v2 consists of an investigation of additional inductive losses for unseen classes along with a semantic guided discriminator.  Empirically, we show consistently  that CIZSL losses can improve generative ZSL models on the challenging task of generalized ZSL from a noisy text on CUB and NABirds datasets. We also show the advantage of our approach to Attribute-based ZSL on AwA2, aPY, and SUN datasets. We also show that CIZSL-v2 has an  improved performance compared to CIZSL-v1. 
\end{abstract}

\begin{IEEEkeywords}
Zero-shot learning, creativity, unseen classes understanding, generative models, attributes, Vision and Language
\end{IEEEkeywords}}

\maketitle

\IEEEdisplaynontitleabstractindextext


%
\IEEEpeerreviewmaketitle

\IEEEraisesectionheading{\section{Introduction}\label{sec:introduction}}
\IEEEPARstart{W}{ith} hundreds of thousands of object categories in the real world and countless undiscovered species, it becomes unfeasible to maintain hundreds of examples per class to fuel the training needs of most existing recognition systems. 
Zipf's law, named after George Zipf (1902$-$1950), suggests that for the vast majority of the world-scale classes, only a few examples are available for training, validated earlier in language (e.g., ~\cite{zipf1935human,zipf1949human}) and later in vision (e.g., ~\cite{salakhutdinov2011learning}).  This problem becomes even more severe when we target recognition at the fine-grained level. For example, there exist tens of thousands of bird and flower species, but the largest available benchmarks have only a few hundred classes motivating a lot of research on classifying instances of unseen classes,  known as Zero-Shot Learning (ZSL). 

\begin{figure}[!htbp]
	\centering
	\vspace{-3mm}
    \includegraphics[width=8.3cm]{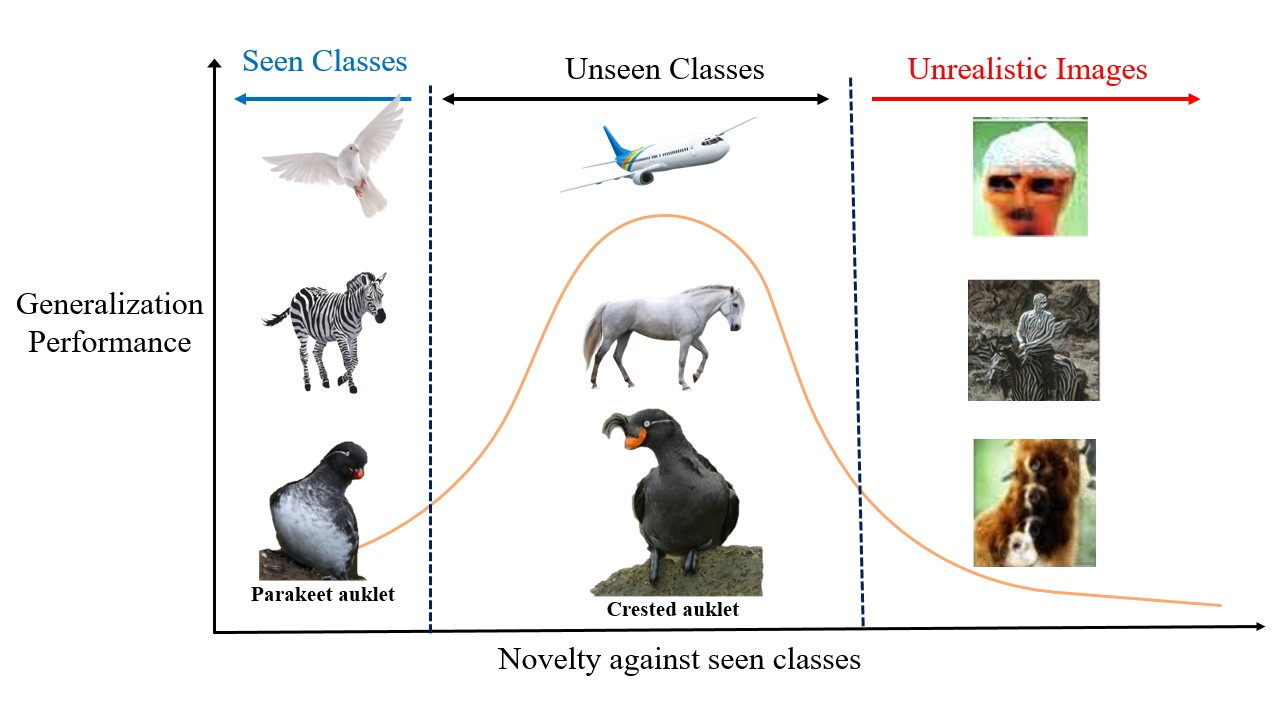}
		\vspace{-3mm}
	\caption{
	Generalizing the learning of zero-shot models requires a deviation from seen classes to accommodate recognizing unseen classes. 
	We carefully model a learning signal that inductively encourages deviation of unseen classes from seen classes, yet not pushed far that the generation falls in the negative hedonic unrealistic range on the right and loses knowledge transfer from seen classes. Interestingly, this curve is similar to the famous Wundt Curve in the Human Creativity literature (Martindale, 1990) ~\cite{martindale1990clockwork}.}
	\label{fig:intro}
\end{figure}

People have a great capability to identify unseen visual classes from text descriptions like
{``The crested auklet is subspecies of birds with dark-gray bodies tails and wings and orange-yellow bill. It is known for its forehead crests, made of black forward-curving feathers.''}; see Fig~\ref{fig:intro} (bottom). We may \emph{imagine} the appearance of ``crested auklet'' in different ways yet all are correct and may collectively help us understand it better.
This  \emph{imagination} notion been modeled in recent ZSL approaches (e.g.,~\cite{guo2017synthesizing, long2017zero, guo2017zero,Elhoseiny_2018_CVPR,xian2018feature}) successfully adopting deep generative models to synthesize visual examples of an unseen object given its semantic description. After training, the model generates imaginary data for each unseen class transforming ZSL into a standard classification task with the generated data. 

However, these generative ZSL methods do not guarantee the discrimination between seen and unseen classes since the generations are not motivated with a learning signal to deviate from seen classes. For example, ``Parakeet Auklet''  as a seen class in Fig~\ref{fig:intro} (left) has a visual text description~\cite{wiki_crested19} that significantly overlaps with ``Crested Auklet'' description, yet one can identify ``Crested Auklet'''s unique ``'black forward-curving feathers'' against ``Parakeet Auklet'' from the text. 
\emph{The core of our work is to address the question of how to produce discriminative generations of unseen visual classes from text descriptions by explicitly learning to deviate from seen classes while allowing transfer to unseen classes.}  
Imagine conditional visual generations' space from class descriptions on an intensity map where light regions imply seen, and darker regions indicate unseen.  These class descriptions are represented in a shared space between the unseen (dark) and the seen (light) classes, and hence the transfer is expected.  This transfer signal is formulated in existing methods by encouraging the generator to produce quality examples conditioned only on the descriptions of the seen classes ( light regions only). In this inductive zero-shot learning, class descriptions of unseen classes are not available during training. They hence can not be used as a learning signal to encourage discrimination across unseen and seen classes explicitly. \emph{Explicitly modeling an inductive and discriminative learning signal from the dark unseen space is at the heart of our work.}

We propose to extend generative zero-shot learning with a discriminative learning signal inspired by the psychology of human creativity. 
Colin Marindale~\cite{martindale1990clockwork}  proposes a psychological theory to explain the perception of human creativity. The definition relates the likability of an art piece to novelty by {\it ``the principle of least effort''}. The aesthetic appeal of artwork first increases when it deviates from existing work till some point, then decreases when the deviation goes too far. This means that it gets difficult to connect this art to what we are familiar with, hence deeming it hard to understand and appreciate. The Wundt Curve can visualize this principle. The X-axis represents novelty, and Y-axis represents likability like an inverted U-shape, similar to the curve in Fig~\ref{fig:intro}. 
We relate the Wundt curve behavior in producing creative art to a desirable generalized ZSL model that has a better capability to distinguish the ``crested auklet'' unseen class from the ``parakeet auklet'' seen class given how similar they are as mentioned before; see Fig~\ref{fig:intro}. A generative ZSL model that cannot deviate generations of unseen classes from instances of seen classes is expected to underperform in generalized zero-shot recognition due to confusion; see Fig~\ref{fig:intro}(left). As the deviation capability increases, the performance is expected to get better but similarly would decrease when the deviation goes too far, producing unrealistic generation and reducing the needed knowledge transfer from seen classes; see Fig~\ref{fig:intro}(middle and right). Our key question is how to properly formulate deviation from generating features similar to existing classes while balancing the desirable transfer learning signal.

\noindent \textbf{Contributions.} 
{\noindent \textbf{1)} We propose a  zero-shot learning approach that explicitly models generating unseen classes by learning to deviate from seen classes carefully. We examine a parametrized entropy measure to facilitate learning how to deviate from seen classes. The psychology of human creativity inspires our approach; thus, we name it Creativity Inspired Zero-shot Learning (CIZSL), and we proposed two versions CIZSL-v1 and an improved version, CIZSL-v2. }

\noindent \textbf{2)} CIZSL-v1 and CIZSL-v2 as creativity inspired models are unsupervised and orthogonal to any Generative ZSL approach.  They can be integrated with any GZSL while neither adding extra parameters nor requiring any additional labels.

\noindent \textbf{3)} By means of extensive experiments on seven benchmarks encompassing Wikipedia-based and attribute-based descriptions, our proposed CIZSL-v1 and CIZSL-v2 can significantly improve the baseline performance.

The source code for CIZSL-v1 is available at  \url{https://github.com/mhelhoseiny/CIZSL}, and the source code for CIZSL-v2 is available at \url{https://github.com/Vision-CAIR/CIZSLv2}.
\section{Related Work}
\noindent\textbf{Early Zero-Shot Learning(ZSL) Approaches} 
 A key idea to facilitate zero-shot learning is finding a common semantic representation that both seen and unseen classes can share. Attributes and text descriptions are shown to be effective shared semantic representations that allow transferring knowledge from seen to unseen classes.
 Lampert \emph{et al.}~\cite{lampert2009} proposed a Direct Attribute Prediction (DAP) model that assumed independence of attributes and estimated the posterior of the test class by combining the attribute prediction probabilities. A parallelly developed, yet similar model was developed by Farhadi \emph{et al.}~\cite{farhadi2009describing}.
 
 \noindent \textbf{Visual-Semantic Embedding ZSL. }Relaxing the unrealistic independence assumption, Akata \emph{et al.}~\cite{akata2016label} proposed an Attribute Label Embedding(ALE) approach that models zero-shot learning as a linear joint visual-semantic embedding. In principle, this model is similar to prior existing approaches that learn a mapping function from visual space to semantic space~\cite{zhang2016learning, shigeto2015ridge}. This has been also investigated in the opposite direction~\cite{zhang2016learning, shigeto2015ridge} as well as jointly learning a function for each space that map to a common space~\cite{yang2014unified, lei2015predicting, akata2015evaluation, romera2015embarrassingly,socher2013zero,Elhoseiny_2017_CVPR,akata2016multi,long2017learning,long2017describing,tsai2017learning, narayan2020latent}. Besides, Shaoteng \emph{et al.}~\cite{liu2020hyperbolic} embedded the visual features into the hyperbolic space to better understand the hierarchical relationships among classes. Guo-Sen \emph{et al.}\cite{xie2020region} incorporated the region-based relation reasoning part into ZSL to capture relationships between different local regions. Further, Wenjia \emph{et al.}~\cite{xu2020attribute} proposed to learn discriminative global and local features by using class-level attributes.


\noindent\textbf{Generative ZSL Approaches}
The notion of generating artificial examples has been recently proposed to model zero-shot learning reducing it to a conventional classification problem~\cite{guo2017synthesizing, long2017zero, guo2017zero,Elhoseiny_2018_CVPR}. Earlier approaches assumed a Gaussian distribution prior for visual space to every class, and the probability densities for unseen classes are modeled as a linear combination of seen class distributions~\cite{guo2017synthesizing}.  Long \emph{et al.}~\cite{long2017zero} instead proposed a one-to-one mapping approach where synthesized examples are restricted.  Recently, Zhu~\emph{et al.}~\cite{Elhoseiny_2018_CVPR},  Xian~\emph{et al.}~\cite{xian2018feature}, and Verma~\emph{et al.}\cite{kumar2018generalized} relaxed this assumption and built on top of  generative adversarial networks (GANs)~\cite{goodfellow2014generative,radford2015unsupervised} to generate examples from unseen class descriptions. Different from ACGAN~\cite{odena2016conditional}, Zhu \emph{et al.} added a visual pivot regularizer (VPG) that encourages generations of each class to be close to the average of its corresponding real features. Maunil \emph{et al.}~\cite{vyas2020leveraging} investigated the relationship between seen classes and unseen classes to guide the generation of the generator.


\noindent\textbf{Semantic Representations in ZSL (e.g., Attributes, Description).}
By definition, ZSL requires additional information (e.g., semantic description of unseen classes) to enable their recognition. A considerable progress has been made in studying attribute representation~\cite{lampert2009, Lampert2014, akata2016label,frome2013devise, zhang2016learning, yang2014unified, lei2015predicting, akata2015evaluation, romera2015embarrassingly, akata2016multi}. Attributes are a collection of semantic characteristics that are filled to describe unseen classes uniquely. Another ZSL trend is to use online textual descriptions ~\cite{elhoseiny2013write,Elhoseiny_2017_CVPR, Qiao2016, reed2016learning, lei2015predicting}. 
Textual descriptions can be easily extracted from online sources like Wikipedia with minimal overhead, avoiding the need to define hundreds of attributes and filling them for each class/image. 
Elhoseiny~\emph{et al.}~\cite{elhoseiny2013write} proposed an early approach for Wikipedia-based zero-shot learning that combines domain transfer and regression to predict visual classifiers from a TF-IDF textual representation~\cite{salton1988term}. Qiao~\emph{et al.}~\cite{Qiao2016} proposed to suppress the noise in the Wikipedia articles by encouraging sparsity of the neural weights to the text terms. Recently, a part-based zero-shot learning model~\cite{Elhoseiny_2017_CVPR} was proposed with a capability to connect text terms to its relevant parts of objects without part-text annotations. 
More recently, Zhu~\emph{et al.}~\cite{Elhoseiny_2018_CVPR} showed that suppressing the non-visual information is possible by the predictive power of their model to synthesize visual features from the noisy Wikipedia text. 
Our work also focuses on the challenging task of recognizing objects based on Wikipedia articles and is also a generative model. Unlike existing, we explicitly model the careful deviation of unseen class generations from seen classes.







\noindent \textbf{Visual Creativity.} 
Computational Creativity studies building machines that generate original items with realistic and aesthetic characteristics~\cite{machado2000nevar,mordvintsev2015inceptionism,dipaola2009incorporating}. Although GANs~\cite{goodfellow2014generative,radford2015unsupervised,ha2017neural} is a powerful generative model, yet it is not explicitly trained to create novel content beyond the training data. For instance, a GAN model trained on artworks might generate the ``Mona Lisa'' again, but would not produce a novel content that it did not see. It is not different for some existing style transfer work~\cite{Gatys2016ImageStyleTransfer,dumoulin2016learned} since there is no incentive in these models to generate new content. More recent work 
adopts computational creativity literature to create novel art and fashion designs~\cite{elgammal2017can,sbai2018design}.
Inspired by~\cite{martindale1990clockwork},   Elgammal et.atl., ~\cite{elgammal2017can} adapted GANs to generate unconditional creative content (paintings) by encouraging the model to deviate from existing painting styles. 
Fashion is a 2.5 trillion dollar industry and has an impact on our everyday life; this motivated~\cite{sbai2018design} to develop a model that can, for example, create an unseen fashion shape ``pants to extended arm sleeves''. 
These models' key idea is to add novelty loss that encourages the model to explore image generation's creative space. 

\section{Background}\label{sec_bg}
\vspace{-2mm}  
GANs~\cite{goodfellow2014generative,radford2015unsupervised} train the generator $G$, with parameters $\theta_G$, to produce samples that the Discriminator $D$ believe they are real. On the other hand, the Discriminator $D$, with parameters $\theta_D$, is trained to classify samples from the real distribution $p_{data}$ as real (1), and samples produced by the generator as fake (0); see Eq~\ref{GAN_org}.
\begin{equation}
\vspace{-5mm}
\small
  \min_{\theta_G} \mathcal{L}_{G} = \min_{\theta_G} \sum_{z_i\in \mathbb{R}^n} \log (1-D(G(z_i))) \\
\end{equation}
\vspace{1em}
\begin{equation}
\small
  \min_{\theta_D} \mathcal{L}_{D} =\min_{\theta_D} \sum_{\substack{x_i}\in\mathcal{D}, z_i\in \mathbb{R}^n} - \log D(x_i) - \log (1-D(G(z_i))))
    \label{GAN_org}
\end{equation}
where $z_i$ is a noise vector sampled from prior distribution $p_{z}$ and $x$ is a real sample from the data distribution $p_{data}$.
In order to learn to deviate from seen painting styles or fashion shapes, \cite{elgammal2017can,sbai2018design} proposed an additional head for the discriminator $D$ that predicts the class of an image (painting style or shape class). During training, the Discriminator $D$ is trained to predict the class of the real data through its additional head, apart from the original real/fake loss.  The generator $G$ is then trained to generate examples that are not only classified as real but more importantly are encouraged to be hard to classify using the additional discriminator head. More concretely, 
\begin{equation}
\small 
  \mathcal{L}_{G} = \mathcal{L}_{G\mbox{ \scriptsize{real/fake}}} + \lambda  \mathcal{L}_{G\mbox{ \scriptsize{creativity}}} 
\end{equation}
The common objective between~\cite{elgammal2017can} and ~\cite{sbai2018design} is to produce novel generations with high entropy distribution over existing classes but they are different in the loss function. 
In~\cite{elgammal2017can}, $\mathcal{L}_{G\mbox{ \scriptsize{creativity}}}$ is defined as the binary cross entropy (BCE) over each painting style produced by the discriminator additional head and the uniform distribution (i.e., $\frac{1}{K}$, $K$ is the number of classes). Hence, this loss is a summation of BCE losses over all the classes. In contrast, Sbai~\emph{et al.}~\cite{sbai2018design} adopted the Multiclass Cross Entropy (MCE) between the distribution over existing classes and the uniform distribution. To our knowledge, creative generation has not been explored before conditioned on text and to also facilitate recognizing unseen classe, \emph{two key differences to our work.} 
{
Relating computational creativity to zero-shot learning is one of the novel aspects in our work by encouraging the deviation of generative models from seen classes. However, proper design of the learning signal is critical to (1) hallucinate class text-descriptions whose visual generations can help the careful deviation, (2) allow discriminative generation while allowing transfer between seen and unseen classes to facilitate zero-shot learning.
}

\section{CIZSL-v1}
\label{sec_cizslv1}
\begin{figure*}[t!]
\vspace{-3mm}
	\centering
	\includegraphics[width=15.0cm,  height=6cm]{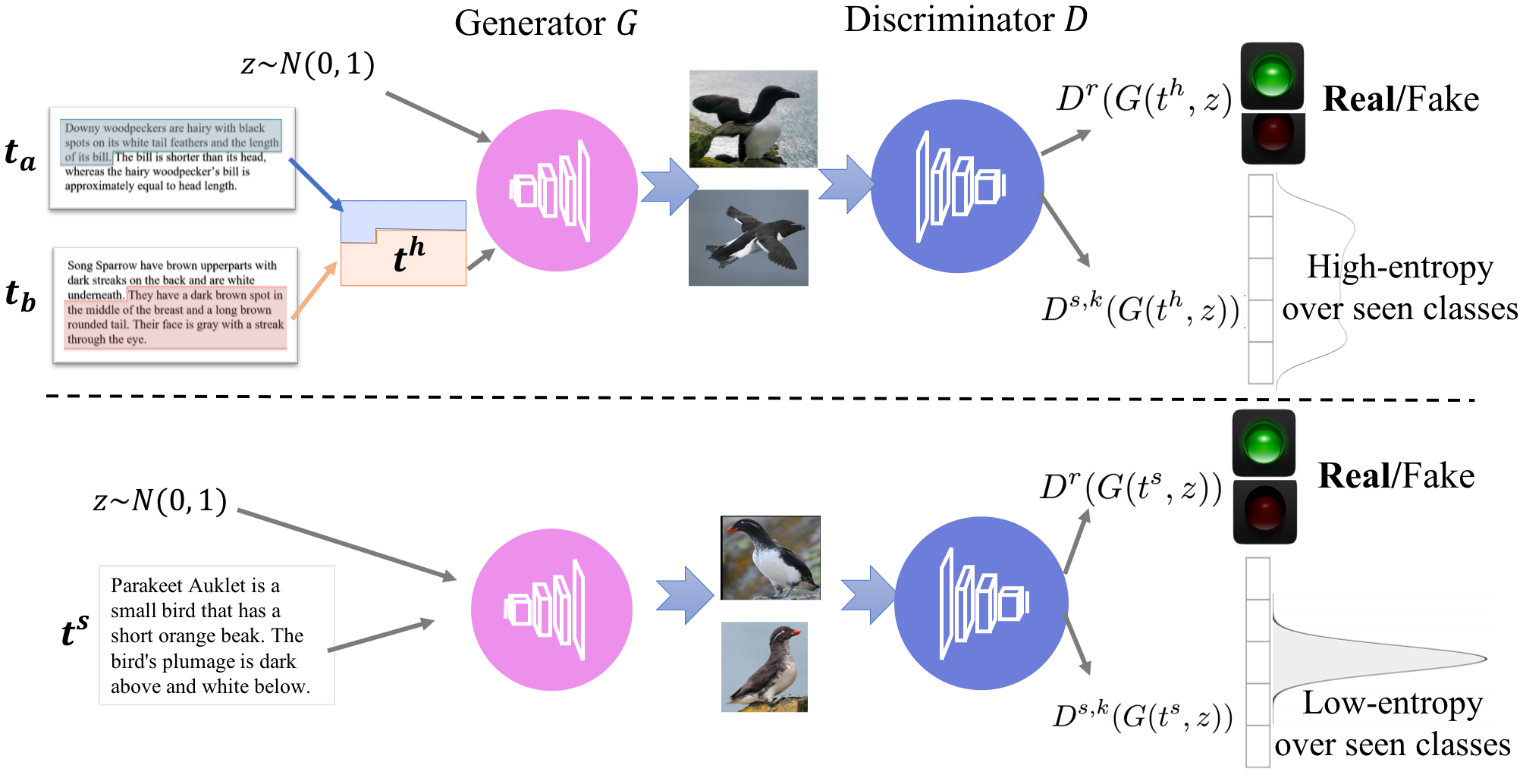}
	\vspace{-3mm}
	\caption{Generator $G$ is trained to carefully deviate from seen to unseen classes without synthesizing unrealstic images.  Top part: $G$ is provided with a hallucinated text $t^h$ and trained to trick discriminator to believe it is real, yet it encourages to deviate learning from seen classes by maximizing entropy over seen classes given $t^h$. Bottom part: $G$ is provided with text of a seen class $t^s$ and is trained to trick discriminator to believe it is real with a corresponding class label(low-entropy).}
	\label{fig:framework}
    \vspace{-5mm}
\end{figure*}

\noindent \textbf{Problem Definition.} We start by defining the zero-shot learning setting. 
We denote the semantic representations of unseen classes and seen classes as $t_i^u=\phi(T_k^u) \in \mathcal{T}$ and $t_i^s \in \mathcal{T}$ respectively, where $\mathcal{T}$ is the semantic space (e.g., features $\phi(\cdot)$ of a Wikipedia article $T_k^u$).  Let's denote the seen data as $D^s = \{(x_i^s, y_i^s, t_i^s) \}^{N^s}_{i=1}$, where $N^s$ is the number of training(seen) image examples,  where $x_i^s \in \mathcal{X}$ denotes the visual features of the $i^{th}$ image in the visual space $\mathcal{X}$, $y_i^s$ is the corresponding category label.  We denote the number of unique seen class labels as $K^s$.  We denote the set of seen and unseen class labels as $\mathcal{S}$ and $\mathcal{U}$, where  the aforementioned $y_i^s \in \mathcal{S}$. Note that the seen and the unseen classes  are disjointed, i.e.,  $\mathcal{S} \cap \mathcal{U} = \emptyset$. 
For unseen classes, we are given their semantic representations, one per class, $\{t_i^u\}_{i=1}^{K^u}$, where $K^u$ is the number of unseen classes. The zero-shot learning (ZSL) task is to predict the label $y_u \in \mathcal{U}$ of an  unseen class visual example  $x^u \in \mathcal{X} $. In the more challenging Generalized ZSL (GZSL),  the aim is to predict $y \in \mathcal{U} \cup \mathcal{S}$ given  $x$ that may belong to seen or unseen classes. 

\noindent  \textbf{Approach Overview.} Fig.~\ref{fig:framework} shows an overview of our Creativity Inspired Zero-Shot Learning model(CIZSL).  Our method builds on top of GANs~\cite{goodfellow2014generative} while conditioning on semantic representation from raw Wikipedia text describing unseen classes. We denote the generator as $G$:  $\mathbb{R}^Z \times \mathbb{R}^T \xrightarrow{\theta_G} \mathbb{R}^X$ and  the discriminator as $D$ : $\mathbb{R}^X \xrightarrow{\theta_D} \{0,1\} \times \mathbb{L}_{cls}$, where  $\theta_G$ and $\theta_D$ are parameters of the generator and the discriminator as  respectively, $\mathbb{L}_{cls}$ is the set of seen class labels (i.e., $\mathcal{S} = \{ 1 \cdots K^s\}$). For the Generator $G$ and as in ~\cite{xian2018feature}, the text representation is then concatenated with a random vector $z \in \mathbb{R}^Z$ sampled from Gaussian distribution $\mathcal{N} (0,1)$; see Fig.~\ref{fig:framework}.  In the architecture of ~\cite{Elhoseiny_2018_CVPR}, the encoded text $t_k$ is first fed to a fully connected layer to reduce the dimensionality and to  suppress the noise  before concatenation with $z$. 
In our work, the discriminator $D$ is trained not only to predict real for images from the training images and fake for generated ones but also to identify the input image category.  
We denote the  real/fake probability produced by $D$ for an input image as $D^r(\cdot)$, and the classification score of a seen class $k \in \mathcal{S}$  given the image as $D^{s,k}(\cdot)$. Hence, the features are generated from the encoded text description $t_k$, as follows $ \tilde{x}_k \leftarrow G(t_k, z)$. The discriminator then has two heads. The first head is an FC layer for binary real/fake classification. The second head is a $K^s$-way classifier over the seen classes. 
Once our generator is trained, it is then used to hallucinate fake generations for unseen classes, where a conventional classifier could be trained as we detail later in Sec~\ref{prediction}.

The generator $G$ is the key imagination component that we aim to train to generalize to unseen classes guided by signals from the discriminator $D$. In Sec~\ref{sec_cizsl}, we detail the definition of our Creativity Inspired Zero-shot Signal to augment and improve the learning capability of the generator $G$. In Sec~\ref{creativegeneration}, we show how our proposed loss can be easily integrated into adversarial generative training.

\subsection{Creativity  Inspired  Zero-Shot  Loss (CIZSL) }\label{sec_cizsl}
We explicitly explore the unseen/creative space of the generator $G$ with a  hallucinated text ($t^h\sim p^h_{text}$). We define $p^h_{text}$, as a probability distribution over hallucinated text description that is likely to be unseen and hard negatives to seen classes. To sample $t^h \sim p^h_{text}$, we first pick two seen text features at random $t^s_a, t^s_b \in \mathcal{S}$. Then we sample $t^h$ by interpolating between them as
\begin{equation}
\small 
  t^h= \alpha t^s_a + (1-\alpha) t^s_b 
  \label{eq_th}
\end{equation}
where $\alpha$ is uniformally sampled between $0.2$ and $0.8$. We discard $\alpha$ values close to $0$ or $1$ to avoid sampling a text feature very close to a seen one.  We also tried different ways to sample $\alpha$ which modifies $p^h_{text}$ like fixed $\alpha=0.5$ or $\alpha\sim \mathcal{N}(\mu=0.5, \sigma=0.5/3)$ but we found uniformally sampling from 0.2 to 0.8 is simple yet effective.

We define our \emph{creativity inspired zero-shot loss }$L_G^C$ based on $G(t^h,z)$ as follows
\begin{equation}
\begin{aligned}
    L_G^C = - \mathbb{E}_{z \sim {p_{z}, t^h \sim {p^s_{text}}}}[D^r(G(t^h, z))] \,\,\,\,\,\,\,\,\,\,\,\,\,\,\,\, \\
    + \mathcolor{black}{\lambda \mathbb{E}_{z \sim {p_{z}, t^h \sim {p^h_{text}}}} [ L_{e}( \{ D^{s,k}(G({t^h}, z))\}_{ k=1 \to K^s })]}  
\end{aligned}
\label{eq_lgc}
\end{equation}
We encourage $G(t^h,z)$ to be real (first term) yet hard to classify to any of the seen classes (second term) and hence achieve more discrimination against seen classes; see Fig.~\ref{fig:framework} (top). More concretely, the first term encourage the generations given $t^h \sim p^h_{text}$ to trick the discriminator to believe it is real (i.e., maximize $D^r(G(t^h, z)$). This loss encourages the generated examples to stay realistic while deviating from seen classes. In the second term, we quantify the classification difficulty by maximizing an entropy function $L_{e}$ that we define later in this section. Minimizing $L_G^C$ connects to the principle of least effort by  Martindale \emph{et al}. 1990, where exaggerated novelty would decrease the transferability from seen classes (see visualized in Fig.~\ref{fig:intro}). Promoting the aforementioned high entropy distribution incents discriminative generation. However, it does not disable knowledge transfer from seen classes since the unseen generations are encouraged to be an entropic combination of seen classes. 
We did not model deviation from seen classes as an additional class with label $K^s+1$ that we always classify $G(t^h,z)$ to, since this reduces the knowledge transfer from seen classes as we demonstrate in our results. 

\noindent \textbf{Definition of $L_e$ :} 
$L_e$ is defined over the seen classes' probabilities, produced by the second discriminator head $\{D^{s,k}(\cdot)\}_{k=1 \to K^s}$ (i.e.,  the softmax output over the seen classes). We tried different entropy maximization losses. They are based on minimizing the divergence between the softmax distribution produced by the discriminator given the hallucinated text features and the uniform distribution. Concretely, the divergence, also known as relative entropy, is minimized between $\{ D^{s,k}(G({t^h}, z))\}_{ k=1 \to K^s }))$ and $\{\frac{1}{K^s}\}_{ k=1 \to K^s }$; see Eq~\ref{le_eq}. 
Note that similar losses have been studied in the context of the creative visual generation of art and fashion(e.g., ~\cite{elgammal2017can,sbai2018design}). However, the focus there was mainly unconditional generation, and there was no need to hallucinate the input text  $t^h$ to the generator, which is necessary in our case; see Sec~\ref{sec_bg}. In contrast, our work also relates two different modalities (i.e., Wikipedia text and images).

\begin{equation}
\small 
\begin{aligned}
L^{KL}_{e} = \sum_{k=1}^{K^s} \frac{1}{K^s} {{D^{s,k}(G(t^h,z))}} \,\,\,\,\,\,\,\,\,\,\,\,\,\,\,\,\,\,\,\,\,\,\,\,\,\,\,\,\,\,\,\,\,\,\,\,\,\,\,\,\,\,\,\,\,\,\,\,\,\,\,\,\,\,\,\,\,\,\,\,\,\,\,\,\,\,\,\,\,\,\,\,\,\,\,\,\,\,\,\\
L^{SM}_{e}(\gamma,\beta) =\frac{1}{\beta-1} \left[ \sum_{k=1}^{K^s} (D^{s,k}(G(t^h,z)^{1-\gamma} {(\frac{1}{K^s})}^{\gamma})^\frac{1-\beta}{1-\gamma} -1\right]
\end{aligned}
\label{le_eq}
\end{equation}

Several divergence/entropy measures has been proposed in the information theory literature~\cite{Renyi60,Tsallis88,Bhattacharyya,IsSM07,sm_1976}. We adopted two divergence losses, the well-known Kullback-Leibler(KL) divergence in $L^{KL}_{e}$ and the two-parameter Sharma-Mittal(SM)~\cite{sm_1976} divergence in $L^{SM}_{e}$ which is relatively less known; see Eq~\ref{le_eq}. It was shown in~\cite{IsSM07},  that other divergence measures are special case of Sharma-Mittal(SM) divergence by setting its two parameters $\gamma$ and $\beta$. It is equivalent to  \emph{R\'enyi}~\cite{Renyi60} when $\beta \to 1$ (single-parameter), \emph{Tsallis} divergence~\cite{Tsallis88} when $\gamma=\beta$ (single-parameter), Bhattacharyya divergence when$\beta \to 0.5$ and $\gamma \to 0.5$,  and \emph{KL} divergence when $\beta \to 1$ and $\gamma \to 1$ (no-parameter). So, when we implement SM  loss, we can also  minimize any of the aforementioned special-case  measures; see details in Appendix A. Note that we also learn $\gamma$ and $\beta$ when we train our model with SM loss.

\subsection{Integrating CIZSL in Adversarial Training} 
\label{creativegeneration}
The integration of our approach is simple that $L_G^C$ defined in Eq~\ref{eq_lgc} is just added to the generator loss; see Eq~\ref{eq:gen}. Similar to existing methods, when the generator $G$ is provided with  text describing a seen class $t^s$, its is trained to trick the discriminator to believe it is real and to predict the corresponding class label (low-entropy for  $t^s$ versus hig-entropy for $t^h$); see Fig~\ref{fig:framework}(bottom). Note that the remaining terms, that we detail here for concreteness of our method, are simlar to existing generative ZSL approaches~\cite{xian2018feature,Elhoseiny_2018_CVPR} 

\noindent\textbf{Generator Loss}
\label{sec_gen_loss}
The generator loss is an addition of four terms, defined as follows
\vspace{-0.5em}
\begin{equation}
\small 
\begin{aligned}
L_G = { L_G^C} - \mathbb{E}_{z \sim {p_{z}, (t^s,y^s) \sim {p^s_{text}}}}[D^r(G(t^s, z))+ \,\,\,\,\,\,\,\,\,\,\,\,\,\,\,\,\,\,\,\,\,\,\,\,\,\,\,\, \,\,\,\,\,\,\,\,\\ \sum_{k=1}^{K^s} y^s_k log(D^{s,k}(G(t^s, z)))]  \\
+ \frac{1}{K^s}\sum_{k=1}^{K^s}|| \mathbb{E}_{z \sim {p_{z}}}[G(t_k, z)] - \mathbb{E}_{x \sim {p^k_{data}}}[x]||^2 \,\,\,\,\,\,\,\,\,\,\,\,
\label{eq:gen}
\end{aligned}
\end{equation}
The first term is our creativity inspired zero-shot loss $L_G^C$, described in Sec~\ref{sec_cizsl}. 
Note that seen class text descriptions $\{t_k\}_{k=1 \to K^s}$ are encouraged to predict a low entropy distribution since loss is minimized when the corresponding class is predicted with a high probability. Hence, the second term tricks the generator to classify visual generations from seen text $t^s$ as real. The third term encourages the generator to be capable of generating visual features conditioned on a given seen text.  The fourth term is an additional visual pivot regularizer that we adopted from~\cite{Elhoseiny_2018_CVPR}, which encourages the centers of the generated (fake) examples for each class $k$ (i.e., with $G(t_k, z)$) to be close to the centers of real ones from sampled from $p^k_{data}$ for the same class $k$. 
Similar to existing methods, the loss for the discriminator is defined as:
\begin{equation}
\small 
\begin{aligned}
L_D = & \mathbb{E}_{z \sim {p_{z}}, (t^s,y^s) \sim {p^s_{text}}}[D^r(G(t^s, z))] - \mathbb{E}_{x \sim {p_{data}}}[D^r(x)]  \\
&+  L_{Lip} -  \frac{1}{2}  \mathbb{E}_{x,y \sim {p_{data}}}[\sum_{k=1}^{K^s} y_k log(D^{s,k}(x)) ] \\
&-  \frac{1}{2} \mathbb{E}_{z\sim p_z, (t^s,y^s) \sim {p^s_{text}}}[\sum_{k=1}^{K^s} y^s_k log(D^{s,k}(G(t^s, z))) ] \\ 
\end{aligned}
\label{eq:disc}
\end{equation}
where $y$ is a one-hot vector encoding of the seen class label for the sampled image $x$, $t^s$ and $y^s$ are
features of a text description and the corresponding on-hot label sampled from seen classes $p^s_{text}$. The first two terms approximate Wasserstein distance of the distribution of real features and fake features. The third term is the gradient penalty to enforce the Lipschitz constraint: $L_{Lip} =  (||\bigtriangledown_{\tilde{x}} D^r(\tilde{x})||_2 - 1)^2$, where  $\tilde{x}$ is the linear interpolation of the real feature $x$ and the fake feature $\hat{x}$; see~\cite{gulrajani2017improved}.  The last two terms are classification losses of the seen real features and fake features from  text descriptions of seen category labels.
\noindent \textbf{Training.}  We construct two minibatches for training the generator $G$, one from seen class $t^s$ and from the halllucinated text $t^h$ to minimize $L_G$ (Eq.~\ref{eq:gen}) and in particular $L_G^C$ (Eq.~\ref{eq_lgc}).  The generator is optimized to fool the discriminator into believing the generated features as real  either from hallucinated text $t^h$ or the seen text  $t^s$. In the mean time,  we maximize their entropy over the seen classes if the generated features comes from hallucinated text $t^h \sim p^h_{text}$ or to the corresponding class if from a real text $t^s$. Training the discriminator is similar to existing works.

\begin{algorithm}[!htbp]
	\begin{algorithmic}[1]
		\STATE{\bfseries Input:} the maximal loops $N_{step}$, the batch size $m$, the iteration number of discriminator in a loop $n_{d}$, the balancing parameter $\lambda_p$, Adam hyperparameters $\alpha_1$, $\beta_1$, $\beta_2$. 
        \FOR{iter $= 1,..., N_{step}$}
        \STATE Sample random text minibatches $t_a, t_b$, noise $z^h$
        \STATE Construct $t^h$ using Eq.6 with different $\alpha$ for each row in the minibatch
        \STATE $\tilde{x}^h \gets G(t^h, z^h)$
		\FOR{$t = 1$, ..., $n_{d}$}
        \STATE Sample a minibatch of images $x$,  matching texts $t$, random noise $z$
		\STATE $\tilde{x}   \gets G(t, z)$
		\STATE Compute the discriminator loss $L_D$ using Eq. 4
        \STATE $\theta_D \gets \text{Adam}(\bigtriangledown_{\theta_D} L_D, \theta_D, \alpha_1, \beta_1, \beta_2)$
        \ENDFOR
        
        \STATE Sample a minibatch of class labels $c$, matching texts $T_c$, random noise $z$
        \STATE Compute the generator loss $L_G$ using Eq. 5
		\STATE\begin{varwidth}[t]{\linewidth}  $\theta_G \gets \text{Adam}(\bigtriangledown_{\theta_G}  L_G ,	\theta, \alpha_1, \beta_1, \beta_2)$ 	
		\end{varwidth}
		\STATE\begin{varwidth}[t]{\linewidth}  $\theta_E \gets \text{Adam}(\bigtriangledown_{\theta_E}  L_G ,	\theta, \alpha_1, \beta_1, \beta_2)$
		\end{varwidth}
		\ENDFOR
	\end{algorithmic}
	\caption{ Training procedure of our approach. We use default values of $n_{d} =5$, $\alpha = 0.001$, $\beta_1 = 0.5$, $\beta_2 =0.9$}
	\label{alg_can_label}
\end{algorithm}

To train our model, we consider visual-semantic feature pairs, images and text, as a joint observation. Visual features are produced either from real data or synthesized by our generator. 
We illustrate in algorithm~\ref{alg_can_label} how $G$ and $D$ are alternatively optimized with an Adam optimizer. The algorithm summarizes the training procedure. In each iteration, the discriminator is optimized for $n_{d}$ steps (lines $6-11$), and the generator is optimized for $1$ step (lines $12-14$). It is important to mention that when $L_e$ has parameters parameters like $\gamma$ and $\beta$ for Sharma-Mittal(SM) divergence, in Eq.~7, that we update these parameters as well by an Adam optimizer and we perform min-max normalization for $L_e$ within each batch to keep the scale of the loss function the same. We denote the parameters of the entropy function as $\theta_E$ (lines $15$). Also, we perform min-max normalization at the batch level for the entropy loss in equation 5

\subsection{Zero-Shot Recognition Test}
\label{prediction}
After training, the visual features of unseen classes can be synthesized by the generator conditioned on a given unseen text  description $t_u$, as $x_u = G(t_u, z)$. 
We can generate an arbitrary number of generated visual features by sampling different $z$ for the same text $t_u$.
With this synthesized data of unseen classes, the zero-shot recognition becomes a conventional classification problem.  We  used nearest neighbor prediction, which we found simple and  effective.

\section{CIZSL-v2 with Semantic Guided Categorizer}
\label{sec_cizslv2}




\begin{figure}
    \centering
    \includegraphics[width=0.5\textwidth]{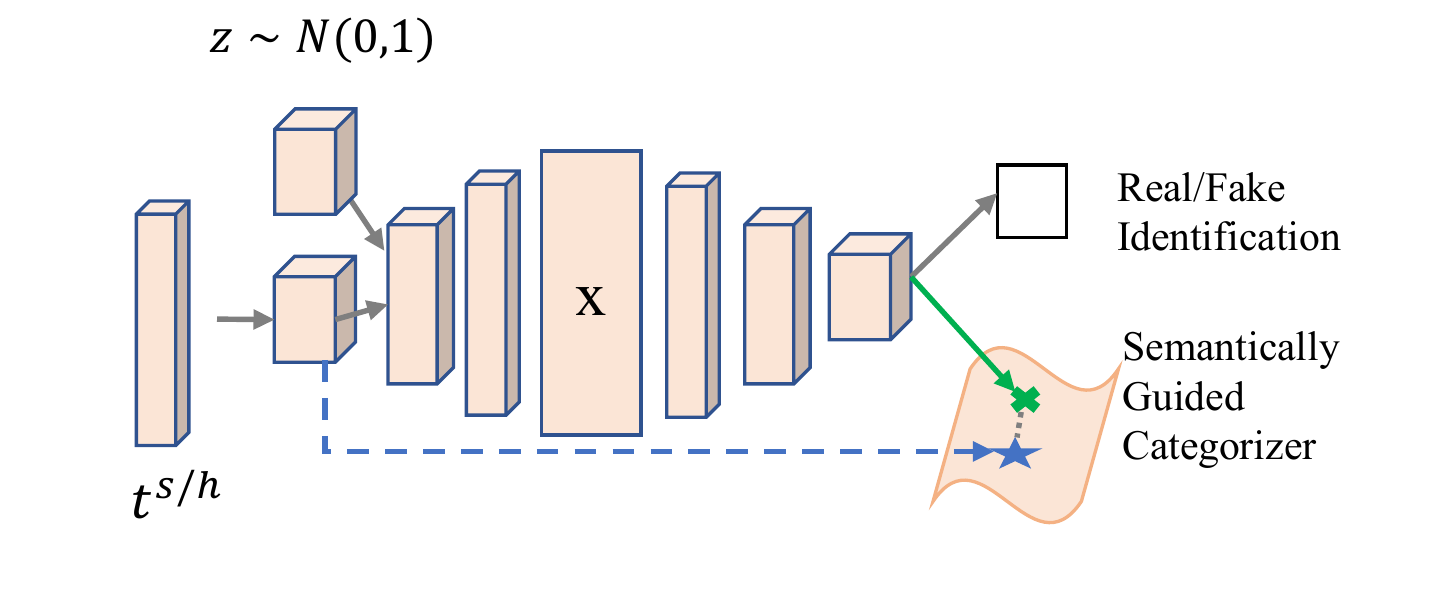}
    \caption{The illustration of the semantic-guided categorizer.}  
    \label{fig:sgc-illustrations}
\end{figure}

 \subsection{Semantic-Guided Categorizer} 
 
 To have a better semantic representation of the generated features, we propose the semantically guided categorizer (SeGC), optimized by the semantic softmax loss. We illustrate our setting at Fig. \ref{fig:sgc-illustrations}. We replace the original classification head of the discriminator with the SeGC, which is computed by the discriminator's final output feature and the reduced dim text feature. We choose the reduced dim text feature instead of the original text feature is because the latter may contain lots of noise, and the features learned by the generator may be more representative. 

The proposed SeGC first compute the extracted visual features to the semantic space and get the compatibility score of the image feature $x^l$ and semantic text feature $t^c$ as follows:
\begin{equation}
    S_c = <x^l W, t_c>
    \label{eq_Sc}
\end{equation}

After instantiating the intput of softmax as the compatibility scores, we obtain the following semantic softmax loss:
\begin{equation}
\begin{split}
L_{Cat} &=-\mathbb{E}_{x \sim p_{g}}[\log p(c \mid x ; C)] \\
&=-\mathbb{E}_{x \sim p_{g}}\left[\log \frac{\exp \left(S_{c}\right)}{\sum_{j \in \mathcal{Y}_{s}} \exp \left(S_{j}\right)}\right]
\end{split}
\end{equation}
To minimize the semantic-guided categorical loss, the generator is encouraged to synthesize more semantic discriminative features.  As we show later, semantic guided discriminator helps improve the ZSL performance in most benchmarks. We think this is the feature spaces that guide the classification decision is driven by the same conditioning signal, making the feature embedding of seen and unseen classes more semantically consistent. 


\subsection{Interpolation and Extrapolation of Hallucinated Text of Unseen Classes}
According to CIZSL-v1,  the hallucinated text $t^h\sim p^h_{text}$ is generated by interpolating between two seen text features $t_a^s, t_b^s\in \mathcal{S}$ randomly by the following equations:
\begin{equation}
    t^{h}=\alpha t_{a}^{\prime}+(1-\alpha) t_{b}^{*}
\end{equation}
Previously, we've tried $\alpha$ uniformly sampled between 0.2 and 0.8, discarding  $\alpha$ values close to 0 or 1 to avoid sampling a text feature very close to a seen one.
In contrast to CIZSL-v1,  we explore extrapolation strategies, which means we sample $\alpha$ less than $-1$ or greater than $+1$ and different combinations. Similar to interpolation, to prevent sampling closer to text features from seen classes, we ignore the values near $-1$ and $+1$. {We generate hallucinated text in different ways to evaluate whether extrapolation strategy may help}. This goes beyond the semantic space that is covered by convex combinations of existing seen classes.  

\section{Experiments}


We investigate the performance of our approach on two class-level semantic settings: textual  and attribute descriptions. Since the textual based ZSL is a harder problem, we used it to run an ablation study for  zero-shot retrieval and generalized ZSL. Then, we conducted  experiments for both settings to validate the generality of our work.

\noindent \textbf{Cross-Validation} 
The weight $\lambda$ of our loss in Eq~\ref{eq_lgc} is a hyperparameter  that we found easy to tune on all of our experiments. 
We start by splitting the data into training and validation split with nearly 80-20\% ratio for all settings. Training and validation classes are selected randomly prior to the training. Then, we compute validation performance when training the model on the 80\% split every 100 iterations out of 3000 iterations. We investigate a wide range of values for $\lambda$, and the value that scores highest validation performance is selected to be used at the inference time. Finally, we combine training and validation data and evaluate the performance on testing data.

\noindent \textbf{Zer-Shot Performance Metrics.} We use two metrics widely used in in evaluating ZSL recognition performance: Standard Zero-shot recognition with the Top-1 unseen class accuracy and Seen-Unseen Generalized Zero-shot performance with Area under Seen-Unseen curve~\cite{chao2016empirical}. The Top-1 accuracy is the average percentage   of  images from unseen classes classifying correctly to one of unseen class labels. However, this might be  incomplete measure since it is more realistic at inference time to encounter also seen classes. Therefore, We also report a generalized zero-shot recognition metric with respect to the seen-unseen  curve, proposed by Chao~\emph{et al.}~\cite{chao2016empirical}. This metric classifies images of both seen $\mathcal{S}$ and unseen classes $\mathcal{U}$ at test time. Then, the performance of a ZSL model is assessed by classifying these images to the label space that covers both seen classes and unseen labels $\mathcal{T} = \mathcal{S}\cup\mathcal{U}$. A balancing parameter is used sample seen and unseen class test accuracy-pair. This pair is plotted as the $(x,y)$ co-ordinate to form the Seen-Unseen Curve(SUC). We follow ~\cite{Elhoseiny_2018_CVPR} in using the Area Under SUC to evaluate the generalization capability of class-level text zero-shot recognition, and the haromnic mean of SUC for attribute-based zero-shot recognition. In our model, we use the trained GAN to synthesize the visual features for both training and testing classes.
\begin{table*}
\centering
\begin{tabular}{@{}cccccccccc@{}}
\toprule
Metric & \multicolumn{4}{c}{Top-1 Accuracy (\%)} &  & \multicolumn{4}{c}{Seen-Unseen AUC (\%)} \\ \cmidrule(lr){2-5} \cmidrule(l){7-10} 
Dataset & \multicolumn{2}{c}{CUB} & \multicolumn{2}{c}{NAB} &  & \multicolumn{2}{c}{CUB} & \multicolumn{2}{c}{NAB} \\
Split-Mode & Easy & Hard & Easy & Hard &  & Easy & Hard & Easy & Hard \\ \midrule
\textbf{CIZSL SM-Entropy (ours final)} & \textbf{44.6} & \textbf{14.4} & 36.5 & \textbf{9.3} &  & \textbf{39.2} & \textbf{11.9} & \textbf{24.5} & \textbf{6.4} \\ \hline 
CIZSL SM-Entropy (replace $2^{nd}$ term in Eq~\ref{eq_lgc} by Classifying $t^h$ as new class) & 43.2 & 11.31 & 35.6 & 8.5 &  &38.3 & 9.5 & 21.6 & 5.6\\ 
CIZSL SM-Entropy (minus  $1^{st}$ term in Eq~\ref{eq_lgc}) & 43.4 & 10.1 & 35.2 & 8.3 &  & 35.0 & 8.2 & 20.1  & 5.4\\  
CIZSL SM-Entropy: (minus  $2^{nd}$ term in Eq~\ref{eq_lgc}) & 41.7 & 11.2 & 33.4 & 8.1 &  & 33.3 & 10.1 & 21.3 & 5.1\\
\hline 
CIZSL Bachatera-Entropy ($\gamma=0.5,\beta =0.5$) & 44.1 & 13.7 & 35.9 & 8.9 &  & 38.9 & 10.3 & 24.3 & 6.2\\
CIZSL Renyi-Entropy ($\beta \to 1$) &  44.1 & 13.3 & 35.8 & 8.8 &  & 38.6 & 10.3 & 23.7 & 6.3\\	
CIZSL KL-Entropy ($\gamma \to 1,\beta \to 1$)	&  44.5 & 14.2 & 36.3 & 8.9 &  & 38.9 & 11.6 & 24.3 & 6.2\\
CIZSL Tsallis-Entropy ($\beta = \gamma$) & 44.1 & 13.8 & \textbf{36.7} & 8.9 &  & 38.9 & 11.3 & 24.5 & 6.3  \\	  \hline
CIZSL SM-Entropy: (minus  $1^{st}$ and $2^{nd}$ terms in Eq~\ref{eq_lgc})= GAZSL~\cite{Elhoseiny_2018_CVPR}& 43.7 & 10.3 & 35.6 & 8.6 &  & 35.4 & 8.7 & 20.4 & 5.8 \\ 
\bottomrule
\end{tabular}%
\caption{CIZSL-v1 Ablation Study using Zero-Shot recognition on \textbf{CUB} \& \textbf{NAB} datasets with two split settings each. CIZSL is GAZSL~\cite{Elhoseiny_2018_CVPR}+ our loss}
\label{tb:nab_cub_ablation}
\end{table*}

\begin{figure}
\centering
\includegraphics[width=0.50\textwidth, height=6.2cm]{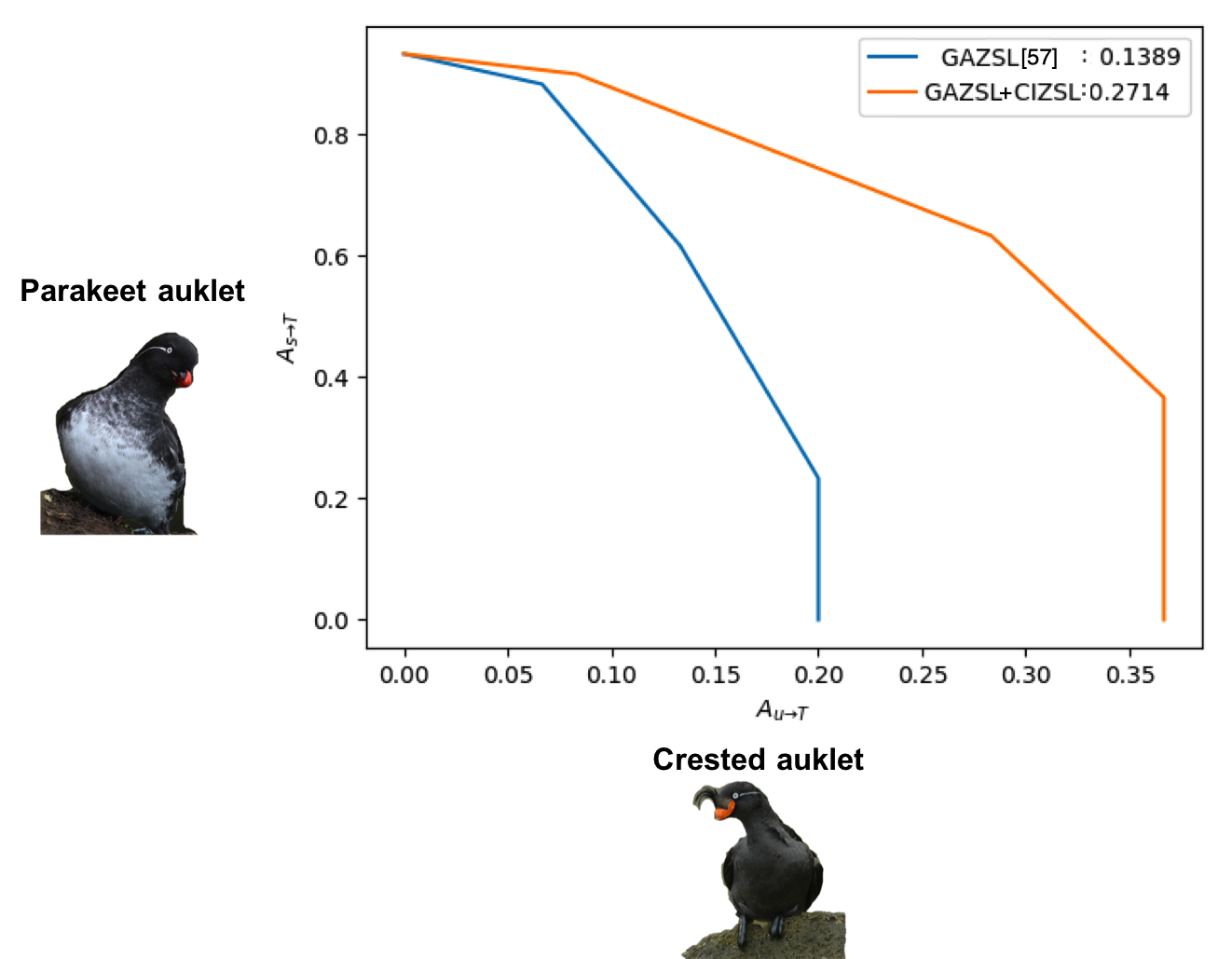}
\captionof{figure}{Seen Unseen Curve for Parakeet Auklet (Seen,  y-axis) vs Crested Auklet (Unseen, x-axis) for GAZSL[57] and     GAZSL[57]+ CIZSL-v1.}
\label{fig_auklet}
\end{figure}

\subsection{Wikipedia based ZSL  Results (4 benchmarks)}
\noindent \textbf{Text Representation.} Textual features for each class are extracted from corresponding raw Wikipedia articles collected by~\cite{elhoseiny2013write,Elhoseiny_2017_CVPR}. 
We used Term Frequency-Inverse Document Frequency (TF-IDF)~\cite{salton1988term} feature vector of dimensionality 7551 for CUB and 13217 for NAB.  

\noindent \textbf{Visual Representation.} We use features of the part-based FC layer in VPDE-net~\cite{zhang2016learning}. The image are fed forward to the VPDE-net after resizing to $224\times 224$, and the feature activation for each detected part is extracted which is of 512 dimensionality. The dimensionalities of visual features for CUB and NAB are 3583 and 3072 respectively. There are six semantic parts shared in CUB and NAB:``head", ``back", ``belly", ``breast", ``leg", ``wing", ``tail". Additionally, CUB has an extra part which is ``leg" which makes its feature representation 512D longer compared to NAB (3583 vs 3072). 
Zhang \emph{et al.}~\cite{zhang2016spda} showed that fine-grained recognition of bird species can be improved by detecting objects parts and learning a part-based learning representations on top. More specifically, ROI pooling is performed on the  detected bird parts (e.g., wing, head) then semantic features are extracted for each part as a representation.  They named  their network Visual Part Detector/Encoder network (VPDE-net) which has VGG~\cite{simonyan2014very} as backbone architecture. We use the VPDE-net as our feature extractor of images for all our experiments on fine-grained bird recognition data sets, so are all the baselines. 

\noindent \textbf{Datasets.} We use two common fine-grained recognition datasets for textual descriptions: \emph{Caltech UCSD Birds-2011} (CUB)~\cite{wah2011caltech} and \emph{North America Birds} (NAB)~\cite{Horn2015}. CUB dataset contains 200 classes of bird species and their Wikipedia textual description constituting a total of 11,788 images. Compared to CUB, NAB is a larger dataset of birds, containing a 1011 classes and 48,562 images.

\noindent \textbf{Splits.} For both datasets, there are two schemes to split the classes into training/testing (in total four benchmarks): Super-Category-Shared (SCS) or \textit{easy} split and Super-Category-Exclusive Splitting (SCE) or \textit{hard} split, proposed in~\cite{Elhoseiny_2017_CVPR}. Those splits represents the similarity of the seen to unseen classes, such that the former represents a higher similarity than the latter. For SCS (easy), unseen classes are deliberately picked such that for every unseen class, there is at least one seen class with the same super-category. Hence, the relevance between seen and unseen classes is very high, deeming the zero-shot recognition and retrieval problems relatively easier. On the other end of the spectrum,  SCE (hard) scheme, the unseen classes do not share the super-categories with the seen classes. Hence, there is lower similarity between the seen and unseen classes making the problem harder to solve. Note that the easy split is more common in literature since it is more Natural yet the deliberately designed hard-split shows the progress when the super category is not seen that we also may expect. 

\begin{table*}
\centering
\begin{tabular}{@{}cccccccccc@{}}
\toprule
Metric & \multicolumn{4}{c}{Top-1 Accuracy (\%)} &  & \multicolumn{4}{c}{Seen-Unseen AUC (\%)} \\ \cmidrule(lr){2-5} \cmidrule(l){7-10} 
Dataset & \multicolumn{2}{c}{CUB} & \multicolumn{2}{c}{NAB} &  & \multicolumn{2}{c}{CUB} & \multicolumn{2}{c}{NAB} \\
Split-Mode & Easy & Hard & Easy & Hard &  & Easy & Hard & Easy & Hard \\ \midrule
WAC-Linear \cite{elhoseiny2013write} & 27.0 & 5.0 & -- & -- &  & 23.9 & 4.9 & 23.5 & -- \\
WAC-Kernel \cite{elhoseiny2016write} & 33.5 & 7.7 & 11.4 & 6.0 &  & 14.7 & 4.4 & 9.3 & 2.3 \\
ESZSL \cite{romera2015embarrassingly} & 28.5 & 7.4 & 24.3 & 6.3 &  & 18.5 & 4.5 & 9.2 & 2.9 \\
ZSLNS \cite{Qiao2016} & 29.1 & 7.3 & 24.5 & 6.8 &  & 14.7 & 4.4 & 9.3 & 2.3 \\
SynC$_{fast}$ \cite{changpinyo2016synthesized} & 28.0 & 8.6 & 18.4 & 3.8 &  & 13.1 & 4.0 & 2.7 & 3.5 \\
ZSLPP \cite{Elhoseiny_2017_CVPR} &  37.2 & 9.7 &30.3 & 8.1 &  & 30.4 & 6.1 & 12.6 & 3.5 \\ \hline
FeatGen~\cite{xian2018feature} &  43.9 & 9.8 &36.2 & 8.7 &  & 34.1 & 7.4 & 21.3 & 5.6 \\
FeatGen\cite{xian2018feature}+ \textbf{CIZSL-v1} &  44.2$\mathcolor{blue}{\mathbf{ ^{+0.3}}}$  & 12.1 $\mathcolor{blue}{\mathbf{ ^{+2.3}}}$  & 36.3 $ \mathcolor{blue}{\mathbf{ ^{+0.1}}}$ & \textbf{9.8} $\mathcolor{blue}{\mathbf{ ^{+1.1}}}$ &  & 37.4 $\mathcolor{blue}{\mathbf{ ^{+2.7}}}$ & 9.8$\mathcolor{blue}{\mathbf{ ^{+2.4}}}$ & 24.7$\mathcolor{blue}{\mathbf{ ^{+3.4}}}$ & 6.2 $\mathcolor{blue}{\mathbf{ ^{+0.6}}}$ \\  \hline
GAZSL~\cite{Elhoseiny_2018_CVPR}& 43.7 & 10.3 & 35.6 & 8.6 &  & 35.4 & 8.7 & 20.4 & 5.8 \\    \midrule
GAZSL~\cite{Elhoseiny_2018_CVPR} + \textbf{CIZSL-v1} &  \textbf{44.6} $\mathcolor{blue}{\mathbf{ ^{+0.9}}}$ & \textbf{14.4} $\mathcolor{blue}{\mathbf{ ^{+4.1}}}$ & \textbf{36.6} $\mathcolor{blue}{\mathbf{ ^{+1.0}}}$ & 9.3 $\mathcolor{blue}{\mathbf{ ^{+0.7}}}$ &  & \textbf{39.2}$\mathcolor{blue}{\mathbf{ ^{+3.8}}}$ & \textbf{11.9}$\mathcolor{blue}{\mathbf{ ^{+3.2}}}$ & \textbf{24.5}$\mathcolor{blue}{\mathbf{ ^{+4.1}}}$ & \textbf{6.4}$\mathcolor{blue}{\mathbf{ ^{+0.6}}}$\\ \bottomrule
\end{tabular}%
\caption{Zero-Shot Recognition on class-level textual description from \textbf{CUB} and \textbf{NAB} datasets with two-split setting.}
\label{tb:nab_cub}
\end{table*}

\begin{table*}
\centering
\begin{tabular}{@{}cccccccc@{}}
\toprule
 & \multicolumn{3}{c}{Top-1 Accuracy(\%)} &  & \multicolumn{3}{c}{Seen-Unseen H} \\ \cmidrule(lr){2-4} \cmidrule(l){6-8} 
 & AwA2 & aPY & SUN &  & AwA2 & aPY & SUN \\ \midrule
DAP~\cite{Lampert2014} & 46.1 & 33.8 & 39.9 &  & -- & 9.0 & 7.2 \\
SSE~\cite{zhang2015zero} & 61.0 & 34.0 & 51.5 &  & 14.8 & 0.4 & 4.0 \\
SJE~\cite{akata2015evaluation} & 61.9 & 35.2 & 53.7 &  & 14.4 & 6.9 & 19.8 \\
LATEM~\cite{xian2016latent} & 55.8 & 35.2 & 55.3 &  & 20.0 & 0.2 & 19.5 \\
ESZSL~\cite{romera2015embarrassingly} & 58.6 & 38.3 & 54.5 &  & 11.0 & 4.6 & 15.8 \\
ALE~\cite{akata2016label} & 62.5 & 39.7 & 58.1 &  & 23.9 & 8.7 & 26.3 \\
CONSE~\cite{norouzi2013zero} & 44.5 & 26.9 & 38.8 &  & 1.0 & -- & 11.6 \\
SYNC~\cite{changpinyo2016synthesized} & 46.6 & 23.9 & 56.3 &  & 18.0 & 13.3 & 13.4 \\
SAE~\cite{kodirov2017semantic} & 54.1 & 8.3 & 40.3 &  & 2.2 & 0.9 & 11.8 \\
DEM~\cite{zhang2016learning} & 67.1 & 35.0 & 61.9 &  & 25.1 & 19.4 & 25.6 \\
DEVISE~\cite{frome2013devise} & 59.7 & 39.8 & 56.5 &  & \textbf{27.8} & 9.2 & 20.9 \\  \hline
GAZSL~\cite{Elhoseiny_2018_CVPR} & 58.9 & 41.1 & 61.3 &  & 15.4 & 24.0 & 26.7 \\ 
GAZSL~\cite{Elhoseiny_2018_CVPR} + \textbf{CIZSL-v1} & \textbf{67.8} $\mathcolor{blue}{\mathbf{ ^{+8.9}}}$ & {42.1} $\mathcolor{blue}{\mathbf{ ^{+1.0}}}$ & {63.7} $\mathcolor{blue}{\mathbf{ ^{+2.4}}}$ &  & $24.6 \mathcolor{blue}{\mathbf{ ^{+9.2}}}$ & {25.7} $\mathcolor{blue}{\mathbf{ ^{+1.7}}}$ & \textbf{27.8} $\mathcolor{blue}{\mathbf{ ^{+1.1}}}$  \\
\hline \hline
FeatGen~\cite{xian2018feature} & 54.3 & 42.6 & 60.8 &  & 17.6 & 21.4 & 24.9 \\
FeatGen~\cite{xian2018feature} + \textbf{CIZSL-v1} & $60.1\mathcolor{blue}{\mathbf{ ^{+5.8}}}$  & $43.8\mathcolor{blue}{\mathbf{ ^{+1.2}}}$ & $59.4\mathcolor{red}{\mathbf{ ^{-0.6}}}$ &  & $19.1\mathcolor{blue}{\mathbf{ ^{+1.5}}}$  & $24.0 \mathcolor{blue}{\mathbf{ ^{+2.6}}}$ & $26.5 \mathcolor{blue}{\mathbf{ ^{+1.6}}}$ \\
\midrule
cycle-(U)WGAN~\cite{felix2018multi} & 56.2 & 44.6 & 60.3 &  & 19.2 & 23.6 & 24.4 \\
cycle-(U)WGAN~\cite{felix2018multi} + \textbf{CIZSL-v1} & $63.6 \mathcolor{blue}{\mathbf{ ^{+7.4}}}$  & \textbf{45.1} $\mathcolor{blue}{\mathbf{ ^{+0.5}}}$  & \textbf{64.2} $\mathcolor{blue}{\mathbf{ ^{+3.9}}}$   &  & $23.9 \mathcolor{blue}{\mathbf{ ^{+4.7}}}$   & \textbf{26.2} $\mathcolor{blue}{\mathbf{ ^{+2.6}}}$   & $27.6 \mathcolor{blue}{\mathbf{ ^{+3.2}}}$  \\
\bottomrule
\end{tabular}%
\caption{Zero-Shot Recognition on class-level attributes of \textbf{AwA2}, \textbf{aPY} and \textbf{SUN} datasets.}
\label{tb:awa2_apy_sun}
\end{table*}

\noindent \textbf{Ablation Study (Table~\ref{tb:nab_cub_ablation}).}  Our loss is composed of two terms shown that encourage the careful  deviation in Eq~\ref{eq_lgc}.  The first term encourages that the generated visual features from the hallucinated text $t^h$ to deceive the discriminator believing it is real, which restricts synthesized visual features to be realistic. The second term maximizes the entropy using a deviation measure.  
In our work, Shama-Mittal(SM) entropy  parameters $\gamma$ and $\beta$ are learnt and hence adapt the corresponding data and split mode to a matching divergence function, leading to the best results especially in the generalized SUAUC metric; see first row in Table~\ref{tb:nab_cub_ablation}. We first investigate the effect of deviating the hallucinated text by classifying it t a new class $K^s+1$, where $K^s$ is the number of the seen classes. We found the performance is significantly worse since the loss would significantly increase indecencies against seen classes and hence reduces seen knowledge transfer to unseen classes; see row 2 in  Table~\ref{tb:nab_cub_ablation}. When we remove the first term (realistic constraints), the performance degrades  especially under the generalized Seen-Unseen AUC metric because generated visual features became unrealistic; see row 3 in  Table~\ref{tb:nab_cub_ablation} (e.g., 39.2\% to 35.0\% AUC drop for CUB Easy and 11.9\%-8.2\% drop for CUB Hard). 
Alternatively, when we remove the second term (entropy), we also observe a significant drop in performance showing that both losses are complementary to each other; see row 4 in  Table~\ref{tb:nab_cub_ablation} (e.g., 39.2\% to 33.5\% AUC drop for CUB Easy and 11.9\%-10.1\% drop for CUB Hard). 
In our ablation,  applying our approach without both terms (our loss) is equivalent to~\cite{Elhoseiny_2018_CVPR}, shown is the last row in Table~\ref{tb:nab_cub_ablation} as one of the least performing baselines. Note that our loss is applicable to other generative ZSL methods as we show in our state-of-the-art comparisons later in this section. 





We also compare different entropy measures to encourage the deviation from the seen classes: \emph{Kullback-Leibler (KL)}, \emph{R\'enyi}~\cite{Renyi60}, \emph{Tsallis}~\cite{Tsallis88}, \emph{Bhattacharyya}~\cite{Bhattacharyya}; see rows 5-8 in Table~\ref{tb:nab_cub_ablation}. All these divergences measure are special cases of  the two parameter ($\gamma$ , $\beta$) Sharma-Mittal(SM)~\cite{sm_1976} divergence that we implemented.
For instance, Renyi~\cite{Renyi60} and Tsallis~\cite{Tsallis88} on the other hand only learns one parameter and achieves comparable yet lower performance.  Bhattacharyya~\cite{Bhattacharyya} and KL have no learnable parameters an achieves lower performance compared to SM.


\begin{figure}
	\centering
	\subfloat[CUB with SCS (easy) split]{\includegraphics[width= 1.6in]{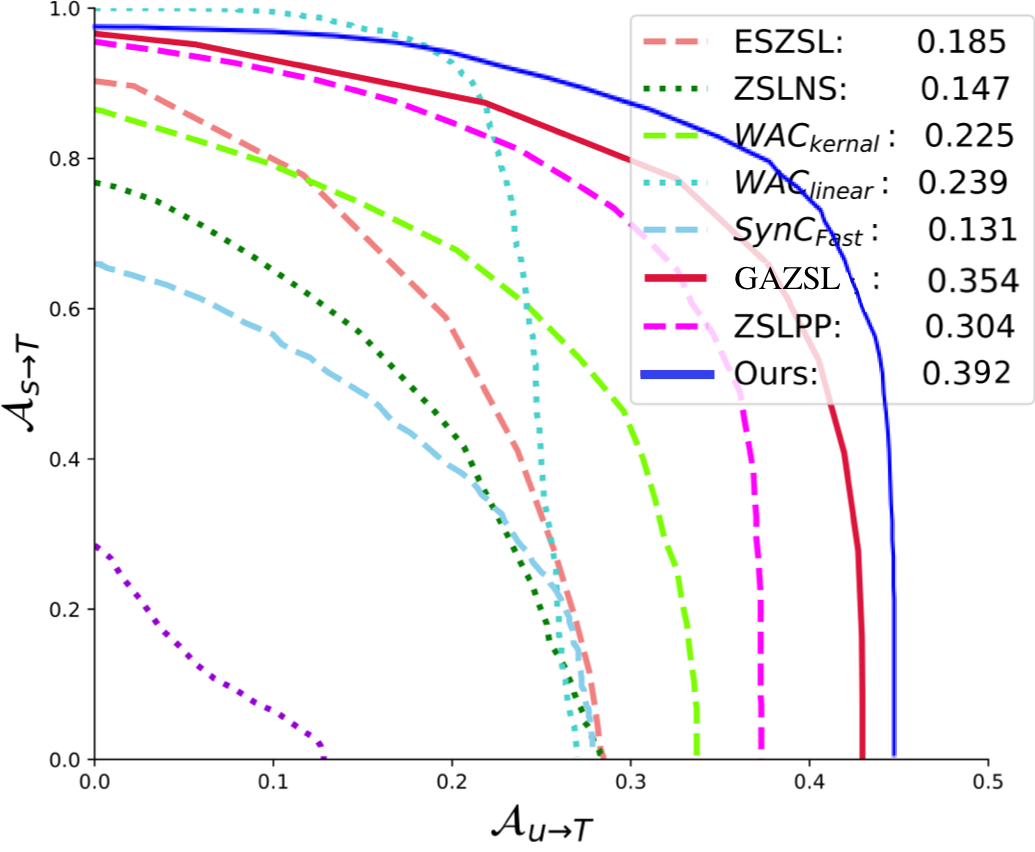}} 
	\subfloat[CUB with SCE (hard) split]{\includegraphics[width= 1.6in]{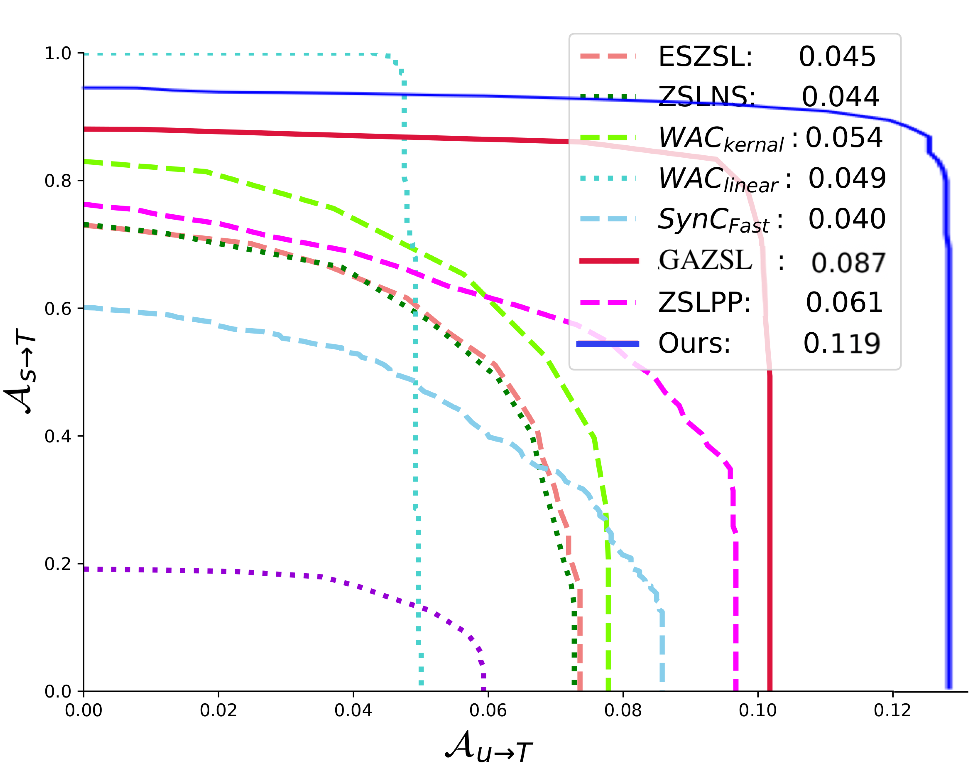}}\\
    \vspace{-2mm}
	\subfloat[NAB with SCS (easy) split]{\includegraphics[width= 1.6in]{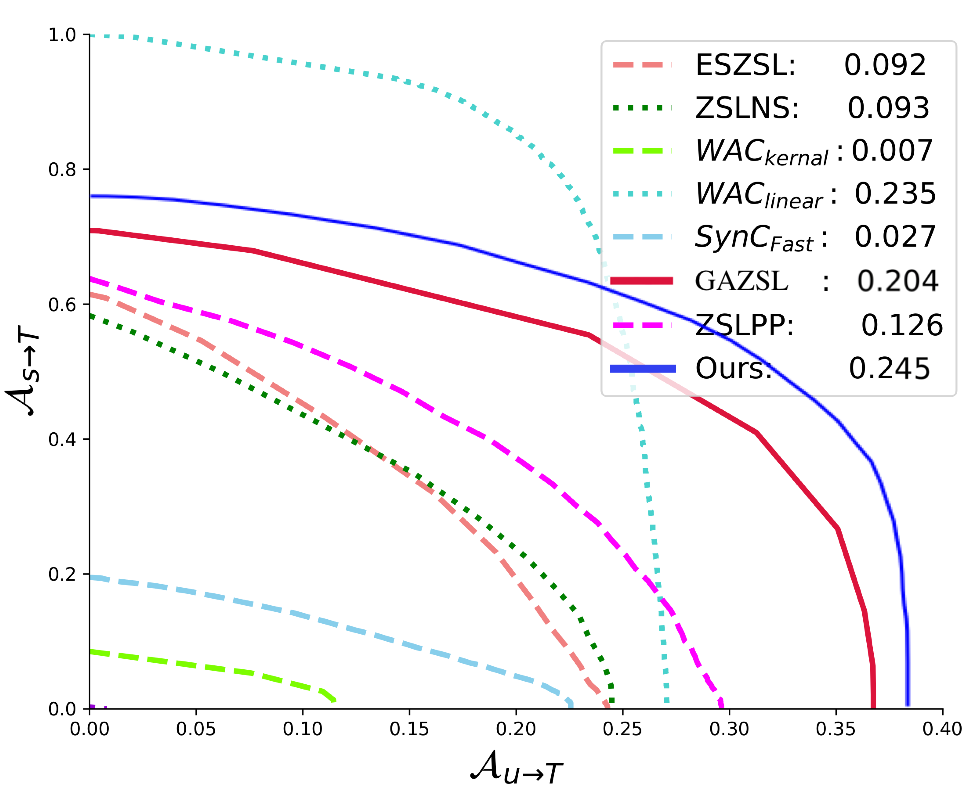}} 
	\subfloat[NAB with SCE (hard) split]{\includegraphics[width= 1.6in]{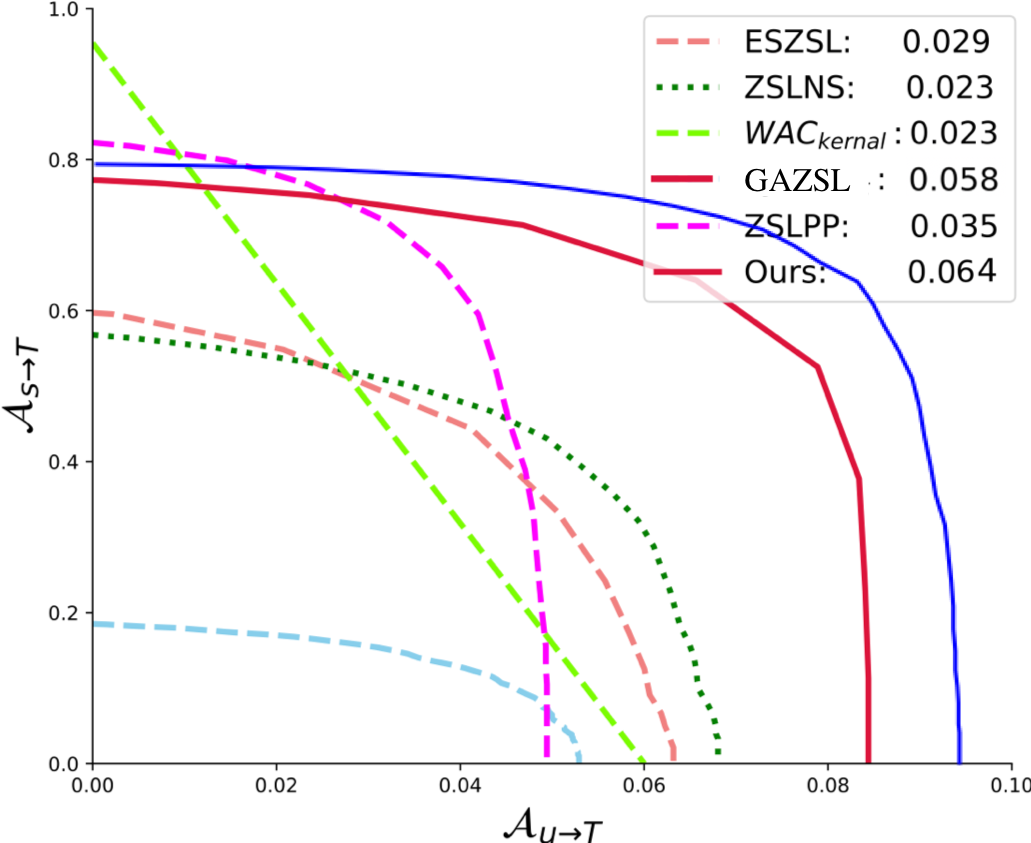}}\\
	\setlength\belowcaptionskip{-2.5ex}
	\caption{Seen-Unseen accuracy Curve with two splits: SCS(easy) and SCE(hard). Ours indicates GAZSL+ CIZSL-v1 }
	\label{fig:GZSL_Curve}
\end{figure}
%

\noindent \textbf{Zero-Shot Recognition and Generality on ~\cite{xian2018feature} and ~\cite{Elhoseiny_2018_CVPR}.}
Fig~\ref{fig_auklet} shows the key advantage of our CIZSL loss,  doubling the capability of [57] from 0.13 AUC to 0.27 AUC  to distinguish between two very similar birds: Parakeet Auklet (Seen class) and Crested Auklet (unseen class), in 200-way classification.  Table~\ref{tb:nab_cub} shows state-of-the-art comparison on CUB and NAB datasets for both their SCS(easy) and SCE(hard) splits (total of four benchmarks). Our method shows a significant advantage compared to the state of the art especially in generalized Seen-Unseen AUC metric ranging from 1.0-4.5\% improvement. Fig~\ref{fig:GZSL_Curve} visualizes Seen-Unseen curves for our four benchmarks CUB (east and hard splits) and NABirds (easy and hard splits) where ours has a significant advantage compared to state-of-the-art on recognizing unseen classes; see our area under SU curve  gain in Fig~\ref{fig:GZSL_Curve}  against the runner-up GAZSL. The average relative SU-AUC improvement on the easy splits is 15.4\% and 23.56\% on the hard split. Meaning, the advantage of our loss becomes more clear as splits get harder, showing a better capability of discriminative knowledge transfer.  We show the generality of our method by embedding it with another feature generation method, FeatGen~\cite{xian2018feature}, causing a consistent improvement.  All the methods are using same text and visual representation.
 \begin{table*}[h!]
\centering
\scalebox{1.0}
{
\begin{tabular}{cccccccc}
\hline
 & \multicolumn{3}{c}{CUB} &  & \multicolumn{3}{c}{NAB} \\ \cline{2-4} \cline{6-8} 
 & 25\% & 50\% & {100\%} &  & 25\% & 50\% &{ {100\% }}\\ \hline
ESZSL \cite{romera2015embarrassingly} & 27.9 & 27.3 & 22.7 &  & 28.9 & 27.8 & 20.9 \\
ZSLNS \cite{Qiao2016} & 29.2 & 29.5 & 23.9 &  & 28.8 & 27.3 & 22.1 \\
ZSLPP \cite{Elhoseiny_2017_CVPR} & 42.3 & 42.0 & 36.3 &  & 36.9 & 35.7 & 31.3 \\
GAZSL \cite{Elhoseiny_2018_CVPR} & 49.7 & 48.3 & 40.3 &  & \textbf{41.6} & 37.8 & 31.0 \\ \hline
GAZSL \cite{Elhoseiny_2018_CVPR}+ CIZSL-v1 & \textbf{50.3}$\mathcolor{blue}{\mathbf{ ^{+0.6}}}$ & \textbf{48.9} $\mathcolor{blue}{\mathbf{ ^{+0.6}}}$ & \textbf{46.2}$\mathcolor{blue}{\mathbf{ ^{+5.9}}}$ &  & 41.0$\mathcolor{red}{\mathbf{ ^{-0.6}}}$& \textbf{40.2}$\mathcolor{blue}{\mathbf{ ^{+2.4}}}$ & \textbf{34.2}$\mathcolor{blue}{\mathbf{ ^{+3.2}}}$\\ \hline
\end{tabular}%
}
\caption{Zero-Shot Retrieval using mean Average Precision(mAP) (\%) on CUB and NAB with SCS(easy) splits.}
\label{tb:Retrieval}
\end{table*}
\noindent \textbf{Zero-Shot Retrieval.} We investigate our model's performance for zero-shot retrieval task given the Wikipedia article of the class using mean Average Precision (mAP), the common retrieval metric. In table ~\ref{tb:Retrieval}, we report the performance of different settings: retrieving 25\%, 50\%, 100\% of the images at each class. 
We follow~\cite{Elhoseiny_2018_CVPR} to obtain the visual center of unseen classes by generating 60 examples for the given text then computing the average. Thus, given the visual center, the aim is to retrieve images based on the nearest neighbor strategy in the visual features space. Our model is the best performing and improves the MAP (100\%) over the  runner-up (~\cite{Elhoseiny_2018_CVPR}) by 14.64\% and 9.61\% on CUB and NAB respectively. 
Even when the model fails to retrieve the exact unseen class, it tends to retrieve visually similar images.




\subsection{Attribute-based Zero-Shot Learning}
\noindent \textbf{Datasets.}  Although it is not our focus, we also investigate the performance of our model's zero-shot recognition ability using different semantic representation. We follow the GBU setting~\cite{gbu}, where images are described by their attributes instead of textual describtion deeming the problem to be relatively easier than textual-description zero-shot learning. We evaluated our approach on the following datasets: Animals with Attributes (AwA2)~\cite{lampert2009}, aPascal/aYahoo objects(aPY)~\cite{farhadi2009describing} and the SUN scene attributes dataset~\cite{patterson2012sun}. They consist of images covering a variety of categories in different scopes: animals, objects and scenes respectively. AwA contains attribute-labelled classes but aPY and SUN datasets have their attribute signature calculated as the average of the instances belonging to each class. 

\noindent \textbf{Zero-Shot Recognition.} 
On AwA2, APY, and SUN datasets, we show in Table~\ref{tb:awa2_apy_sun} that our CIZSL loss improves three generative zero-shot learning models including GAZSL~\cite{Elhoseiny_2018_CVPR},  FeatGen~\cite{xian2018feature},  and  cycle-(U)WGAN~\cite{felix2018multi}.
The table also shows our comparison to the state-of-the-art where we mostly obtain a superior performance. Even when obtaining a slightly lower score than state-of-the-art on AWA2, our loss adds a 9.2\% Seen-Unseen H absolute improvement to the non-creative GAZSL~\cite{Elhoseiny_2018_CVPR}. We also evaluated our loss on \emph{CUB-T1(Attributes) benchmark~\cite{gbu}}, where the Seen-Unseen H for  GAZSL~\cite{Elhoseiny_2018_CVPR}] and GAZSL~\cite{Elhoseiny_2018_CVPR}+CIZSL are 55.8 and  57.4, respectively.

 \section{Analysis}

\begin{figure*}[!htbp]
	\centering
   \includegraphics[width=\textwidth,height=3.8cm]{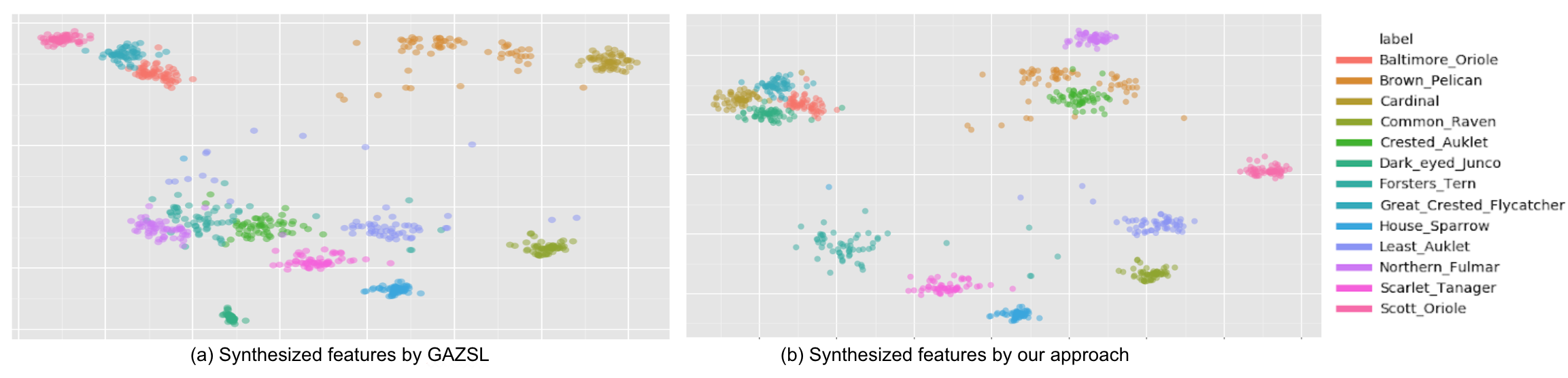}
	\caption{t-SNE visualization of features of randomly selected unseen classes. Compared to GAZSL\cite{Elhoseiny_2018_CVPR}, our  method preserves more inter-class discrimination.}
	\label{fig:tSNE}
\end{figure*}

In Fig.~\ref{fig:tSNE}, we show the t-SNE visualization of the embedding of  13 randomly selected unseen classes with and without CIZSL loss. The results shows that CIZSL loss encouraged the embeddings of different unseen classes to be more distinguishable from each others which improves the zero-shot classification performance. 
In the rest of this section, we perform more analysis and share some negative results that better explain our architecture and loss design choices of CIZSL. 

\subsection{Deeper Network Architectures}
   Generally, there are three different input settings for the original network architecture. For real image features $x$, we feed them into the discriminator, which has one identification head and one classification head. Identification head is used to judge whether the input features of the discriminator is real or fake while the classification head is to classify the image. For the input of text feature $t^s$, we concatenate the features reduced dimensions with a random noize $z$ sampled from gaussian distribution. Then the generator outputs the synthetic image feature $G(t^s, z)$. For hallucinated text feature $t^h$, which is generated by an interpolation from two text features from seen classes, the pipeline is the same as that for $t^s$ and then we get the synthesized image feature $G(t^h, z)$. The difference is that for $G(t^h, z)$, we have a high-entropy setting, which expects the discriminator doesn't give a very high confidence on a particular class for $t^h$. But for $G(t^s, z)$, on the contrary, we have the low-entropy over seen classes expectation.


We've tried several different architectures to make our network more powerful in the zero-shot learning task. As our inputs for the generator are text features $t^s$ or $t^h$, a convolutional layer may  not be a natural choice. We select two results at Tab. \ref{tb:doublenet-results}. {Note that the presented CIZSL-v2 is an updated version of the previous one. We've chosen an improved set of cross-validated hyper-parameters. Our further experiments are based on the new setting.} In DoubleNet,  we double the number of hidden layers both for the generator and the discriminator compared to the original architecture~\cite{elhoseiny2019creativity}. Considering the fact that doubled hidden layers may largely increase the number of training parameters, we also show the results of reduced dims version of DoubleNet.

\begin{table}[b!]
\centering
\resizebox{0.5\textwidth}{!}{%
\begin{tabular}{@{}cccccccccc@{}}
\toprule
Metric & \multicolumn{4}{c}{Top-1 Accuracy (\%)} &  & \multicolumn{4}{c}{Seen-Unseen AUC (\%)} \\ \cmidrule(lr){2-5} \cmidrule(l){7-10} 
Dataset & \multicolumn{2}{c}{CUB} & \multicolumn{2}{c}{NAB} &  & \multicolumn{2}{c}{CUB} & \multicolumn{2}{c}{NAB} \\
Split-Mode & Easy & Hard & Easy & Hard &  & Easy & Hard & Easy & Hard \\ \midrule
\text{GAZSL\cite{Elhoseiny_2018_CVPR} + \textbf{CIZSL-v2}} & 
\textbf{41.9}  & \textbf{15.5} & \textbf{35.0} & \textbf{9.6} & & \textbf{39.2} &  12.6 & \textbf{23.6} & \textbf{6.9}\\ \midrule
DoubleNet & 41.4 & 14.3 & 33.4 & 9.1 & & 38.5 & \textbf{13.1} & 22.0 & 6.6\\ \midrule
DoubleNet-ReducedDims & 41.5 & 14.7 & 33.1 & 8.9 & & 38.3 & 12.9 & 21.2 & 6.2\\ \bottomrule
\end{tabular}%
}
\caption{Zero-Shot Recognition on class-level textual description from \textbf{CUB} and \textbf{NAB} datasets with two-split setting with different architectures.}
\label{tb:doublenet-results}
\end{table} 

The results at Tab. \ref{tb:doublenet-results} are validated with three trials with different creativity weight selected from ${0.0001, 0.001, 0.01, 0.1, 1}$. Our original shallower  model seems to be better than DoubleNet architecture. We performed experiments with deeper architecture but the experiments failed, suggesting  that maybe simpler architecture works better as they are less vulnerable to overfitting. 

\subsection{Interpolation and Extrapolation of Hallucinated Text of Unseen Classes}



\begin{table}[b!]
\centering
\resizebox{0.5\textwidth}{!}{%
\begin{tabular}{@{}cccccccccc@{}}
\toprule
Metric & \multicolumn{4}{c}{Top-1 Accuracy (\%)} &  & \multicolumn{4}{c}{Seen-Unseen AUC (\%)} \\ \cmidrule(lr){2-5} \cmidrule(l){7-10} 
Dataset & \multicolumn{2}{c}{CUB} & \multicolumn{2}{c}{NAB} &  & \multicolumn{2}{c}{CUB} & \multicolumn{2}{c}{NAB} \\
Split-Mode & Easy & Hard & Easy & Hard &  & Easy & Hard & Easy & Hard \\ \midrule
GAZSL~\cite{Elhoseiny_2018_CVPR} + \textbf{CIZSL-v2}\\ \cmidrule{1-1}
$\mathcal{U}(0.2, 0.8)$ & 41.9 & 15.5 & 35.0 & \textbf{9.6} & & 39.2 & 12.6 & \textbf{23.6} & \textbf{6.9}\\ 
$\mathcal{U}(-0.5, -0.2)$ & 43.4 & 15.2 & \textbf{35.3} & 9.4 & & 38.9 & \textbf{12.7} & 23.2 & 6.7\\ 
$\mathcal{U}(1.2, 1.5)$ & 43.5 & 14.8 & 35.0 & 9.5 & & 39.2 & 12.6 & 23.4& 6.8\\ 
$\mathcal{U}(-0.5, -0.2)$ U $(1.2, 1.5)$ & \textbf{43.9} & 15.3 & \textbf{35.3} & 9.7 & & 39.0 & \textbf{12.7} & 23.4 & 6.8\\ 
$\mathcal{U}(-0.5, -0.2)$ U $(0.2, 0.8)$ U $(1.2, 1.5)$ & 43.5 & \textbf{15.6} & 35.2 & 9.5 & & \textbf{39.4} & 12.5 & 23.4 & 6.7\\

\bottomrule
\end{tabular}%
}
\caption{Interpolation and extrapolation of hallucinated text. }
\label{tb:extrapolate-fusion-all}
\end{table}  

We show the results of different interpolation and extrapolation strategies at Fig.~\ref{tb:extrapolate-fusion-all}. By comparing \textit{interpolation only} ($\alpha \in \mathcal{U}(0.2, 0.8)$) with \textit{extrapolations} ($\alpha \in \mathcal{U}(-0.5, -0.2)$ U $(0.2, 0.8)$ U $(1.2, 1.5)$), we can see the extrapolating strategies tend to have better top-1 accuracy on easy tasks while interpolating tends to have better top-1 accuracy on hard ones. As for seen-unseen AUC, there isn't a very huge difference. Besides, we can also find the combination of positive extrapolation and negative extrapolation can lead to better results compared with adopt one only. Further, the combination of all settings will trade-off the performance of interpolation and extrapolation. But in general, there is no huge difference among different hallucination strategies.

\subsection{Adding Creativity Entropy Loss for Hallucinated Text to both Discriminator and Generator}
\label{sec7_dev_desc}



In the original discriminator loss, we've only considered about the discriminative ability with real x and generated features given text features of seen classes. However, what's missing here is the discriminative loss of visual generations of halllucinated text of unseen classes. In other words, there is no motivation for the discriminator to learn from hallucinated text. 
Therefore, we firstly integrate the creativity loss term for the generator to our discriminator hoping our discriminator to not give a very high confidence of a particular class if the input is hallucinated text $t^h$. We show the results at Table. \ref{tb:disc-loss-generator}. The validated hyperparameters for CUB\&NAB easy\&hard are 1, 0.0001, 0.1, 0.1, respectively.





\begin{table}[!htbp]
\centering
\resizebox{0.5\textwidth}{!}{%
\begin{tabular}{@{}cccccccccc@{}}
\toprule
Metric & \multicolumn{4}{c}{Top-1 Accuracy (\%)} &  & \multicolumn{4}{c}{Seen-Unseen AUC (\%)} \\ \cmidrule(lr){2-5} \cmidrule(l){7-10} 
Dataset & \multicolumn{2}{c}{CUB} & \multicolumn{2}{c}{NAB} &  & \multicolumn{2}{c}{CUB} & \multicolumn{2}{c}{NAB} \\
Split-Mode & Easy & Hard & Easy & Hard &  & Easy & Hard & Easy & Hard \\ \midrule

GAZSL~\cite{Elhoseiny_2018_CVPR} + CIZSL-v2 on $G$ & 41.9 & 15.5 & 35.0 & 9.6 & & 39.2 & 12.6 & 23.6 & 6.9\\ 


GAZSL~\cite{Elhoseiny_2018_CVPR} + CIZSL-v2 on $G$ and $D$ & 39.6 & 10.3 & 28.0 & 6.8 & & 35.2 & 9.8 & 12.4 & 4.8\\ 
\bottomrule
\end{tabular}%
}
\caption{Comparative study of adding identification and entropy loss to the generator. }
\label{tb:disc-loss-generator}
\end{table}   

From the results at Tab. \ref{tb:disc-loss-generator}, adding our creativity inspired loss to the discriminator harms the performance. We think this since adding the generative deviation signal makes more sense to add only to the generator. Adding the creativity inspired deviation signal to the discriminator destabilizing the feature learning process which leads to poor generalization.

 \subsection{Zero-Shot Retrieval Qualitative Samples}
 \begin{figure*}[!htbp]
    \centering
    \includegraphics[width=18.0cm,height=6.5cm]{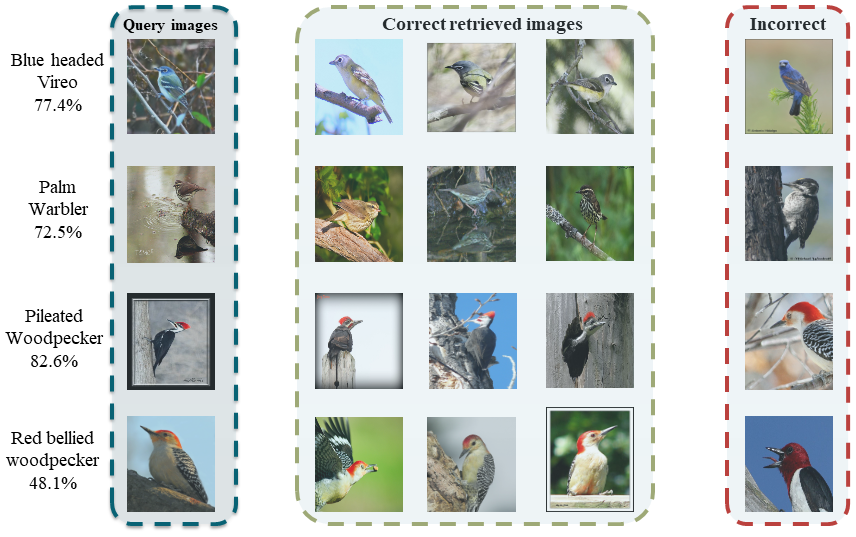}
    \centering
    \includegraphics[width=18.0cm,height=6.5cm]{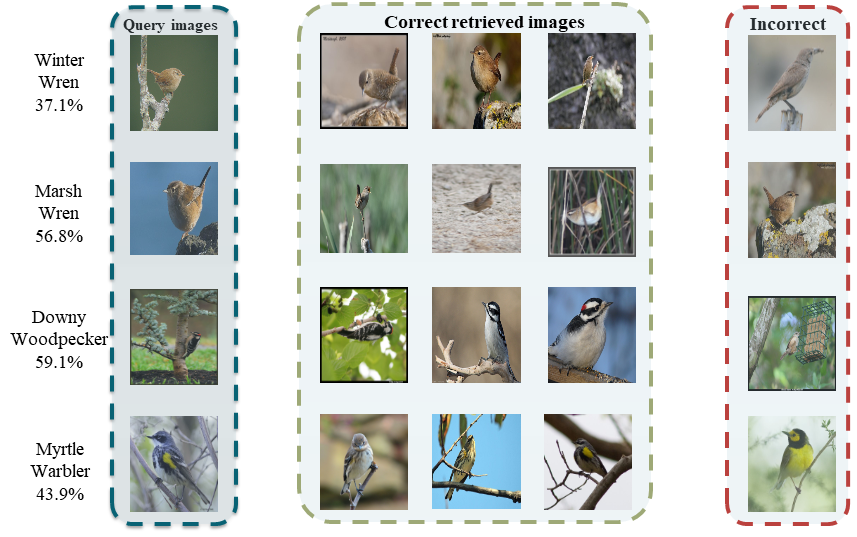}
    \centering
    \includegraphics[width=18.0cm,height=6.5cm]{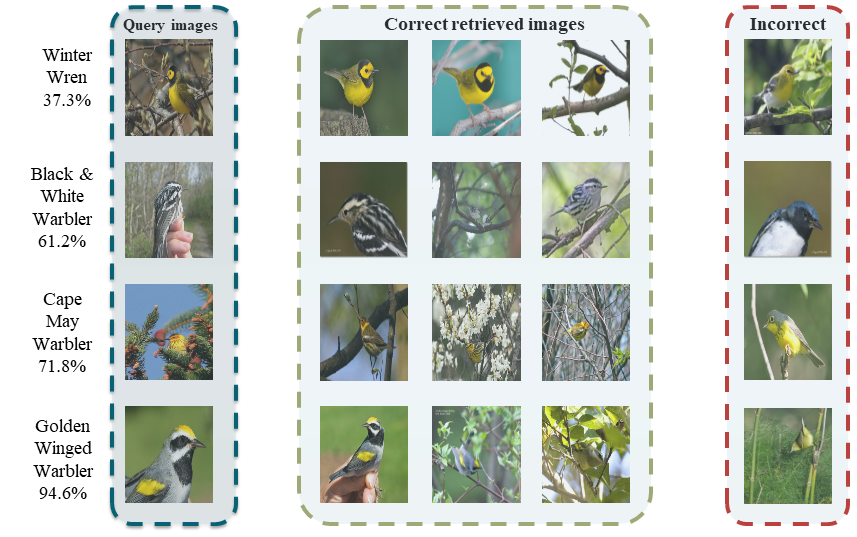}
    \caption{Qualitative results of zero-shot retrieval on CUB dataset using SCS setting.}
    \label{fig_qual}
\end{figure*}
We show in Fig.~\ref{fig_qual} several examples of the retrieval on CUB dataset using SCS split setting. Given a query semantic representation of an unseen class, the task is to retrieve images from this class. Each row is an unseen class. We show three correct retrievals as well as one incorrect retrieval, randomly picked. We note that, even when the method fails to retrieve the correct class, it tends to retrieve visually similar images. For instance, in the Red bellied Woodpecker example (last row in the first subfigure). Our algorithm mistakenly retrieves an image of the red headed woodpecker. It is easy to notice the level of similarity between the two classes, given that both of them are woodpeckers and contain significant red colors on their bodies.
 
\section{Semantic Guided CIZSL-v2 Results}

\subsection{Semantic-Guided Categorizer}

\begin{table}[!htbp]
\centering
\resizebox{0.5\textwidth}{!}{%
\begin{tabular}{@{}cccccccccc@{}}
\toprule
Metric & \multicolumn{4}{c}{Top-1 Accuracy (\%)} &  & \multicolumn{4}{c}{Seen-Unseen AUC (\%)} \\ \cmidrule(lr){2-5} \cmidrule(l){7-10} 
Dataset & \multicolumn{2}{c}{CUB} & \multicolumn{2}{c}{NAB} &  & \multicolumn{2}{c}{CUB} & \multicolumn{2}{c}{NAB} \\
Split-Mode & Easy & Hard & Easy & Hard &  & Easy & Hard & Easy & Hard \\ \midrule

CIZSL-v2 &41.9 & 15.5 &35.0 &9.6 & & 39.2 & 12.6 & 23.6 &6.9\\ 

CIZSL-v2+SeGC & \textbf{42.4} & \textbf{16.4} & \textbf{35.2} & \textbf{10.6} & & \textbf{39.3} & \textbf{14.9} & \textbf{23.8} & \textbf{7.5}\\
\bottomrule
\end{tabular}%
}
\caption{Semantic guided categorizer.}
\label{tb:SeGC-results}
\end{table}   

We show the results of our proposed SeGC at Tab. \ref{tb:sgc-results}. As we can see, CIZSL-v2+SeGC is the semantic-guided categorizer, which means based on CIZSL-v2, we add the SeGC head and remove the original classification head. From the results, we can see the proposed SeGC can  help improve the performance, especially on hard splits.

\subsection{Normalization Scale of Semantic-Guided Categorizer}
\begin{table}[htb!]
\centering
\resizebox{0.5\textwidth}{!}{%
\begin{tabular}{@{}cccccccccc@{}}
\toprule
Metric & \multicolumn{4}{c}{Top-1 Accuracy (\%)} &  & \multicolumn{4}{c}{Seen-Unseen AUC (\%)} \\ \cmidrule(lr){2-5} \cmidrule(l){7-10} 
Dataset & \multicolumn{2}{c}{CUB} & \multicolumn{2}{c}{NAB} &  & \multicolumn{2}{c}{CUB} & \multicolumn{2}{c}{NAB} \\
Split-Mode & Easy & Hard & Easy & Hard &  & Easy & Hard & Easy & Hard \\ \midrule


CIZSL-v2+SeGC \\ \cmidrule{1-1}
Standard (SD) & 42.4 & \textbf{16.4} & 35.2 & \textbf{10.6} & & \textbf{39.3} & \textbf{14.9} & \textbf{23.8} & \textbf{7.5}\\
SD+Norm($\eta=1$) & 42.4 & 15.5 & \textbf{35.4} & 10.0 & & 39.0 & 13.1 & 22.8 & 7.0\\
SD+Norm($\eta=3$) & 42.4 & 15.0 & 34.5 & 9.2 & & 39.1 & 13.2 & 23.3 & 6.7 \\
SD+Norm($\eta=5$) & \textbf{42.5} & 15.1 & 33.7 & 10.3 & & 39.1 & 13.6 & 22.3 & 7.4\\
SD+Norm($\eta=10$) & 42.1 & 15.4 & 33.9 & 9.9 & & 39.2 & 13.3 & 22.3 & 6.8\\
SD+Norm($\eta=20$) & 42.0 & 15.3 & 34.0 & 10.4 & & 38.9 & 13.2 & 22.9 & 7.3\\
\bottomrule
\end{tabular}%
}
\caption{Semantic guided categorizer with normalize-scale.} 
\label{tb:normale-scale}
\end{table}   

To investigate whether the normalization of SeGC can affect the performance, we did performed experiments to explore  the difference among difference normalization scale factors. More concretely, in contrast to Eq.~\ref{eq_Sc}, we L2- normalize $x^l W$ and $t_c$ and multiply each by a scale $\eta$ inspired by~\cite{Bell_2016_CVPR,zhang2019large}; see Eq.~\ref{eq_Sc_NS}.
\begin{equation}
    S_c = \eta^2 \frac{<x^l W, t_c>}{\|x^l W\| \| t_c\|}
    \label{eq_Sc_NS}
\end{equation}
We show the results at Table.~\ref{tb:normale-scale}. According to our current results, it seems the weighted normalization strategy did not improve for our semantically guided categorizer. It performed similarly to SeGC without Normalization and Scaling. {We think this is since the semantic descriptors are already normalized and they are guiding the norm of $\|x^l W\|$ in the case of using SeGC in the discriminator. }

\subsection{Semantically Guided Categorized along with the Real/Fake loss of Hallucinated Text}



{There is no motivation for the discriminator to learn much from hallucinated text based on the current losses does not integrate $t_h$. This is since  entropy loss hurts the performance as we showed earlier in Sec~\ref{sec7_dev_desc}. Here, we decided to incorporate $t^h$ differently by adding   hallucinated real-fake loss to the discriminator as follows.}

\begin{equation}
L_h = \mathrm{E}_{z \sim p_{z}, t_{h} \sim p^{h}}\left[D^{r}\left(G\left(t_{h}, z\right)\right)\right]
\end{equation}

We show the comparative results at Table.~\ref{tb:sgc-results}. The first two rows shows the results  without the semantic guided categorizer. The added hallucinated real-fake loss for the discriminator can improve both the top-1 accuracy and the seen-unseen AUC in this case. 
In the third and the fourth row of  Table.~\ref{tb:sgc-results}, we show results of the semantic guided version with and without the real/fake loss intergrated with the discriminator, denoted as  CIZSL+SeGC and CIZSL+SeGC+R/F. All results were shown based on hyperparameters of CIZSL+SeGC version, 3 trials for CIZSL+SeGC+R/F loss  
In this case, SeGC plus R/F loss strategy is not as helpful compared with the SeGC only version. This shows that semantic guided of the discriminator alleviates the need of the real/fake loss. In other words,  the embedding space is more semantically enriched in this case  which is a key for improving zero-shot learning models. The validated hyperparameters for CUB\&NAB easy\&hard are 1, 0.0001, 0.1, 0.1, respectively. 


\begin{table}[!htbp]
\centering
\resizebox{0.5\textwidth}{!}{%
\begin{tabular}{@{}cccccccccc@{}}
\toprule
Metric & \multicolumn{4}{c}{Top-1 Accuracy (\%)} &  & \multicolumn{4}{c}{Seen-Unseen AUC (\%)} \\ \cmidrule(lr){2-5} \cmidrule(l){7-10} 
Dataset & \multicolumn{2}{c}{CUB} & \multicolumn{2}{c}{NAB} &  & \multicolumn{2}{c}{CUB} & \multicolumn{2}{c}{NAB} \\
Split-Mode & Easy & Hard & Easy & Hard &  & Easy & Hard & Easy & Hard \\ \midrule

CIZSL-v2 &41.9 & 15.5 &35.0 &9.6 & & 39.2 & 12.6 & 23.6 &6.9\\ 

CIZSL-v2+R/F Loss for $t^h$ & \textbf{42.8} & 15.9 & \textbf{35.5} & \textbf{10.6} & & 39.0 & 13.4 & 23.2 & 7.3\\
CIZSL-v2+SeGC & 42.4 & \textbf{16.4} & 35.2 & \textbf{10.6} & & \textbf{39.3} & \textbf{14.9} & \textbf{23.8} & \textbf{7.5}\\
CIZSL-v2+SeGC+R/F loss & 42.1 & 15.7 & 34.8 & 10.3 & & 39.1 & \textbf{13.6} & 22.9 & 7.1\\
\bottomrule
\end{tabular}%
}
\caption{Semantic guided categorizer plus the real/fake loss.}
\label{tb:sgc-results}
\end{table}

\subsection{Categorization Loss on Hallucinated Visual Features of Hallucinated Text in a Semantically Guided Way}
\begin{table}[!htbp]
\centering
\resizebox{0.5\textwidth}{!}{%
\begin{tabular}{@{}cccccccccc@{}}
\toprule
Metric & \multicolumn{4}{c}{Top-1 Accuracy (\%)} &  & \multicolumn{4}{c}{Seen-Unseen AUC (\%)} \\ \cmidrule(lr){2-5} \cmidrule(l){7-10} 
Dataset & \multicolumn{2}{c}{CUB} & \multicolumn{2}{c}{NAB} &  & \multicolumn{2}{c}{CUB} & \multicolumn{2}{c}{NAB} \\
Split-Mode & Easy & Hard & Easy & Hard &  & Easy & Hard & Easy & Hard \\ \midrule
$K^u=100 (w/o)$& \textbf{42.3} & 14.8 & 33.5 & \textbf{9.8} & & \textbf{38.6} & 12.4 & \textbf{21.7} & 7.2\\
$K^u=100 (w/)$ & 41.8 & \textbf{15.2} & \textbf{33.7} & 9.5 & & 38.5 & \textbf{12.9} & 21.3 & \textbf{7.3}\\
\bottomrule
\end{tabular}%
}
\caption{Comparative results of adding categorization loss on hallucinated visual features of hallucinated text in a semantically guided way.}
\label{tb:categorization-hall3}
\end{table}

Here, we show the categorization of hallucinated visual features in a semantically guided way. Concretely, we add to the following loss to the generator.
\begin{equation}
   L_G^u = - \mathbb{E}_{z \sim {p_{z}, (t^s,y^s) \sim {p^s_{text}}}}[ \sum_{k=1}^{K^u} y^u_k log(D^{u,k}(G(t^u_k, z)))] 
\end{equation}
where   $K^u$ is the number of halluncinated unseen classes in the current mini-batch.   $D^{u,k}(\cdot, \cdot)$ is semantically guided discriminator on these $K^u$ hallucinated unseen classes. Note that this is different from the semantically guided classification head on seen classes. Since both classification heads are semantically guided, there is no additional weights introduced by having this additional head to integrate this loss.   Table.~\ref{tb:categorization-hall3} show the experiments where $K^u=100$ with and without having the loss added. The results shows that this additional loss does not add a significant gain to the performance but it offers a slight improvement in the hard split where unseen classes are more different than seen classes compared to the easy split.

\section{Conclusion}
We draw an inspiration from the psychology of human creativity to improve the capability of unseen class imagination for zero-shot recognition. We adopted GANs to discrimnatively imagine visual features given a hallucinated text describing an unseen visual class. Thus, our generator learns to synthesize unseen classes from hallucinated texts. Our loss encourages deviating generations of unseen from seen classes by enforcing a high entropy on seen class classification while being realistic. Nonetheless, we ensure the realism of hallucinated text by synthesizing visual features similar to the seen classes to preserve knowledge transfer to unseen classes. Comprehensive evaluation on seven benchmarks shows consistently that CIZSL losses can improve generative zero shot learning models on zero-shot learning and retrieval with class description defined by  Wikipedia articles and attributes.

 \bibliographystyle{unsrt}  
\bibliography{ref}

\clearpage

\begin{IEEEbiography}[\vspace{-4mm}{\includegraphics[width=1.0in,height=1.0in,clip,keepaspectratio]{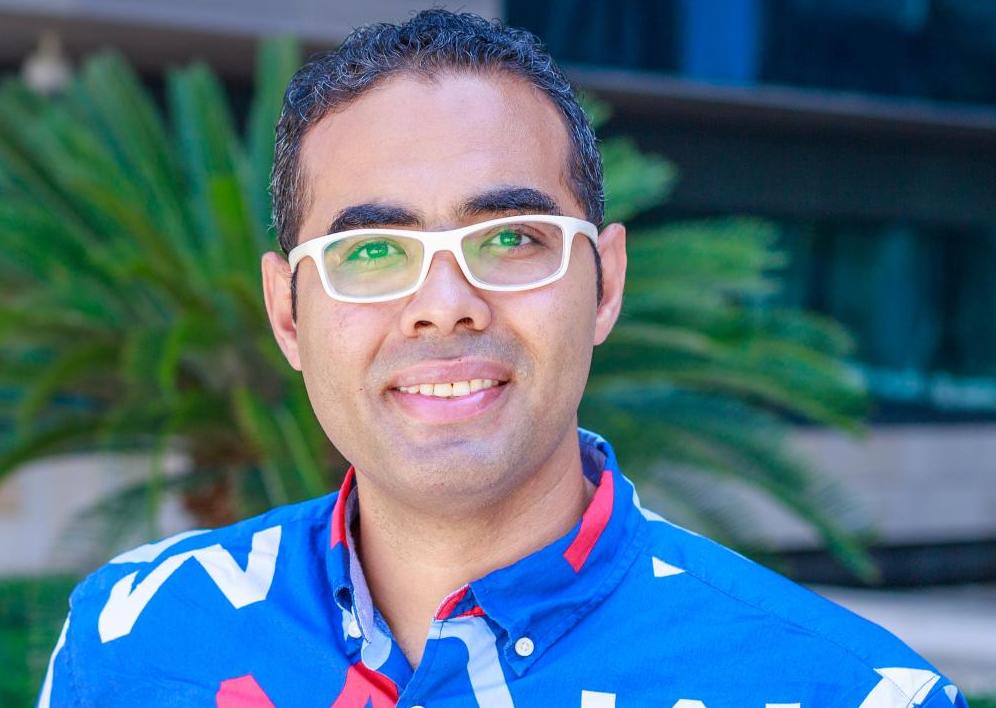}}]{Mohamed Elhoseiny}
 is a professor of Computer Science at KAUST.  Researcher at Facebook Research. Previously, he was a visiting Faculty at Stanford Computer Science department (2019-2020), Visiting Faculty at Baidu Research Silicon Valley Lab (2019), Postdoc researcher at Facebook AI Research (2016-2019). Dr. Elhoseiny did his Ph.D. in 2016 at Rutgers University during which he spent times at Adobe Research (2015-2016) for more than a year and  at SRI International in 2014 (a best intern award). Dr. Elhoseiny received an NSF Fellowship in 2014 for the Write-a-Classifier project (ICCV13) and the Doctoral Consortium award at CVPR 2016. Dr. Elhoseiny  was also a Stanford Igniter as a venture idea generator; Stanford Ignite is an  entrepreneurship program at Stanford Graduate School of Business.
His primary research interest is in computer vision and especially in efficient multimodal learnning with limited data in areas like zero/few-shot learning,  Vision \& Language, language guided visual-perception. He is also interested in Affective AI and especially to produce novel art and fashion with AI. His creative AI work was featured in MIT Tech Review,  New Scientist Magazine, and HBO Silicon Valley. 
\end{IEEEbiography}

\vspace{-10mm}
\begin{IEEEbiography}[\vspace{-4mm}{\includegraphics[width=1.3in,height=1.3in,clip,keepaspectratio]{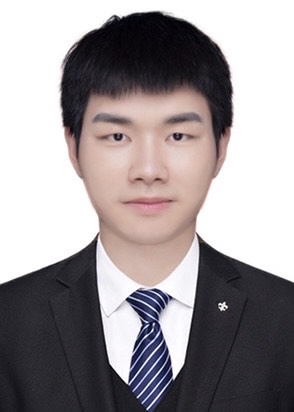}}]{Kai Yi} is a MS/PhD student at the computer Science Department at KAUST supervised by Prof. Elhoseiny.     Kai  received his B.Eng with honor from Department of Software Engineering, Xi'an Jiaotong University in June 2019. His current research interests are computer vision, zero-/few-shot lerning, and computational biology.
Before joining KAUST, he interned at CMU under Prof. Min Xu and NUS supervised by Prof. Angela Yao. Previously, He was a research intern at Sensetime at Wentao Liu's group. Before that, He was fortunate to be an intern with Prof. Nanning Zheng at IAIR (Institute of Artificial Intelligence and Robotics), Prof. Liang Zhao and Prof. Feng Yu at Moral Psycholoy Lab, and Senior Engineer Jing Tao at NEKEY lab (Ministry of Education Key Lab for Intelligent Networks and Network Security.)
\end{IEEEbiography}

\vspace{-10mm}
\begin{IEEEbiography}[\vspace{-5mm}{\includegraphics[width=1.3in,height=1.3in,clip,keepaspectratio]{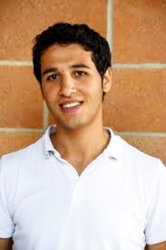}}]{Mohamed  Elfeki}
is a PhD Student in Computer Science department at The University of central Florida. Before Joining UCF, he recieved his bachelor degree from Alexandia University in 2016. He also has spent time as an intern at Facebook AI Research, Amazon Science, and Microsoft Research. Elfeki is  interested in mitigating the fundamental problems of training deep networks on noisy physical-world data and applying them to large-scale real-world problems.
\end{IEEEbiography}

\appendices

\section{Divergence Measures} 


We generalize the expression of the creativity term to a broader family of divergences, unlocking new way of enforcing deviation from seen classes. 

In \cite{IsSM07}, Sharma-Mittal divergence was studied, originally introduced  \cite{SM75}. Given two parameters ($\alpha$ and $\beta$), 
 the Sharma-Mittal (SM) divergence $SM_{\alpha, \beta}(p\|q)$,  between two distributions $p$ and $q$ is defined $\forall \alpha >0,~\alpha \neq 1,~\beta \neq 1$ as 
\begin{equation}
SM(\alpha, \beta)(p || q) = \frac{1}{\beta-1} \left[ \sum_i (p_i^{1-\alpha} {q_i}^{\alpha})^\frac{1-\beta}{1-\alpha} -1\right]
\label{eq:smgen}
\end{equation}
It was shown in~\cite{IsSM07} that most of the widely used divergence measures are special cases of SM divergence. For instance, each of the R\'enyi, Tsallis and Kullback-Leibler (KL) divergences can be defined as limiting cases of SM divergence as follows:
\begin{equation}
\small
\begin{split}
R_{\alpha}(p\|q) = &\lim_{\beta \to 1} {SM_{\alpha, \beta}(p\|q) }  = \frac{1}{\alpha-1} \ln (\sum_i {p_i^\alpha q_i^{1-\alpha} } )),  \\
T_{\alpha}(p\|q) = &\lim_{\beta \to \alpha} {SM_{\alpha, \beta}(p\|q) }= \frac{1}{\alpha-1} (\sum_i {p_i^\alpha q_i^{1-\alpha} } ) -1),  \\
KL(p\|q) = & \lim_{\beta \to 1, \alpha \to 1} {SM_{\alpha, \beta}(p\|q) } = \sum_i{p_i \ln ( \frac{p_i}{q_i}}).
\end{split}
\label{eqdef}
\end{equation}
In particular, the Bhattacharyya divergence ~\cite{bhatt67}, denoted by $B(p\|q)$   is a limit case of SM and R\'enyi divergences as follows as $\beta \to 1, \alpha \to 0.5$
\begin{equation}
\small
\begin{split}
B(p\|q)& = 2 \lim_{\beta \to 1, \alpha \to 0.5}  SM_{\alpha,\beta}(p\|q)  = - \ln \Big( \sum_i p_i^{0.5} q_i^{0.5} \Big).
\end{split}
\end{equation}
Since the notion of creativity in our work is grounded to maximizing the deviation from existing shapes and textures through KL divergence, we can  generalize our MCE creativity loss by minimizing  Sharma Mittal (SM) divergence   between a uniform distribution and the softmax output $\hat{D}$ as follows
\begin{equation}
\begin{split}
\mathcal{L}_{SM} = SM(\alpha,\beta)( \hat{D} || u) =
SM(\alpha, \beta)(\hat{D} || u) \\= \frac{1}{\beta-1} \sum_i (\frac{1}{K}^{1-\alpha} {\hat{D_i}}^{\alpha})^\frac{1-\beta}{1-\alpha} -1
\end{split}
\label{eq_sm}
\end{equation}
\vspace{5mm}

\appendices
\end{document}